\documentclass[10pt,twocolumn,letterpaper]{article}

\usepackage[pagenumbers]{cvpr} % To force page numbers, e.g. for an arXiv version

\usepackage[dvipsnames]{xcolor}

\definecolor{cvprblue}{rgb}{0.21,0.49,0.74}
\usepackage[pagebackref,breaklinks,colorlinks,citecolor=cvprblue]{hyperref}

\usepackage{graphicx}
\usepackage[utf8]{inputenc} % allow utf-8 input
\usepackage{booktabs}       % professional-quality tables
\usepackage{amsfonts}       % blackboard math symbols
\usepackage{nicefrac}       % compact symbols for 1/2, etc.
\usepackage{microtype}      % microtypography
\usepackage{xcolor}         % colors
\usepackage{subcaption}
\usepackage{multirow}
\usepackage{enumitem}
\usepackage{adjustbox}
\usepackage{xspace}
\usepackage{colortbl}
\usepackage{amsmath}
\usepackage{lipsum}
\usepackage{amssymb}
\usepackage{mathtools}
\usepackage{amsthm}
\usepackage{pifont}% http://ctan.org/pkg/pifont
\newcommand{\cmark}{\ding{51}\xspace}%
\newcommand{\xmark}{\ding{55}\xspace}%

\usepackage[toc]{appendix}
\usepackage{etoc}
\usepackage{minitoc}

\definecolor{mygreen}{rgb}{0, 0.6823, 0.7215}
\newcommand{\keyword}[1]{``{#1}''}
\newcommand{\stdv}[1]{{\scriptsize$\pm$#1}}

\def\paperID{13906} % *** Enter the Paper ID here
\def\confName{CVPR}
\def\confYear{2024}

\title{%
Discovering and Mitigating Visual Biases through Keyword Explanation
}

\author{%
Younghyun Kim\thanks{Equal Contribution}$\:\:^1${\quad}Sangwoo Mo$^{*\,2}${\quad}Minkyu Kim$^3${\quad}Kyungmin Lee$^1${\quad}Jaeho Lee$^4${\quad}Jinwoo Shin$^1$\\
$^1$KAIST{\quad}$^2$University of Michigan{\quad}$^3$KRAFTON{\quad}$^4$POSTECH\\
{\small\texttt{younghyun.kim@kaist.ac.kr}{\quad}\texttt{swmo@umich.edu}}
}

\begin{document}
\maketitle

\etocdepthtag.toc{mtchapter}
\etocsettagdepth{mtchapter}{subsection}
\etocsettagdepth{mtappendix}{none}
\faketableofcontents

\begin{abstract}
Addressing biases in computer vision models is crucial for real-world AI deployments. However, mitigating visual biases is challenging due to their unexplainable nature, often identified indirectly through visualization or sample statistics, which necessitates additional human supervision for interpretation. To tackle this issue, we propose the Bias-to-Text (B2T) framework, which interprets visual biases as keywords. Specifically, we extract common keywords from the captions of mispredicted images to identify potential biases in the model. We then validate these keywords by measuring their similarity to the mispredicted images using a vision-language scoring model. The keyword explanation form of visual bias offers several advantages, such as a clear group naming for bias discovery and a natural extension for debiasing using these group names. Our experiments demonstrate that B2T can identify known biases, such as gender bias in CelebA, background bias in Waterbirds, and distribution shifts in ImageNet-R/C. Additionally, B2T uncovers novel biases in larger datasets, such as Dollar Street and ImageNet. For example, we discovered a contextual bias between \keyword{bee} and \keyword{flower} in ImageNet. We also highlight various applications of B2T keywords, including debiased training, CLIP prompting, and model comparison.\footnote{Code: \url{https://github.com/alinlab/b2t}}
\end{abstract}    
\section{Introduction}
\label{sec:intro}

Biased datasets can induce failures in image classifiers, potentially harming model performance and raising fairness concerns~\citep{torralba2011unbiased}. These model failures may manifest as spurious correlations, where specific groups contribute to model errors~\citep{simon1954spurious}, or as distribution shifts, where the test distribution differs from the training distribution~\citep{recht2019imagenet}. 
For instance, in a face dataset, if blond images are predominantly associated with women, the image classifier may misclassify blond faces as women, resulting in fairness issues~\citep{bolukbasi2016man}. Moreover, this bias can impact model performance when evaluated in different scenarios, such as a gender-balanced dataset of blonds~\citep{geirhos2020shortcut}. Therefore, extensive efforts have been devoted to recognizing and addressing biases in models~\citep{mehrabi2021survey,caton2020fairness}.

\begin{figure}[t]
\centering\small
\includegraphics[width=.8\linewidth]{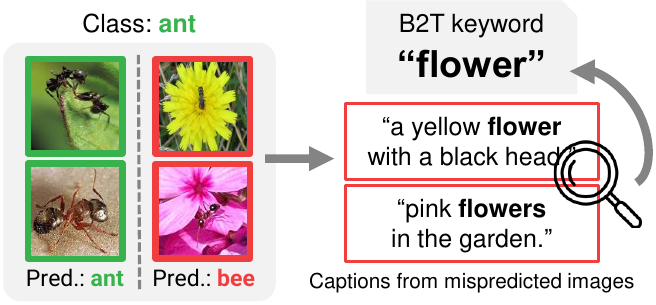}
\vspace{-0.05in}
\caption{
\textbf{Concept.}
Our Bias-to-Text (B2T) framework reveals visual biases of image classifiers in a keyword explanation form. For example, B2T identified novel biases in ImageNet~\cite{deng2009imagenet}. Specifically, the keyword "flower" implies that the classifier associates \keyword{ant} images with \keyword{flower} as \keyword{bees,} indicating contextual bias.
}\label{fig:hook}
\vspace{-0.1in}
\end{figure}
\begin{figure*}[t]
\centering\small
\includegraphics[width=.9\textwidth]{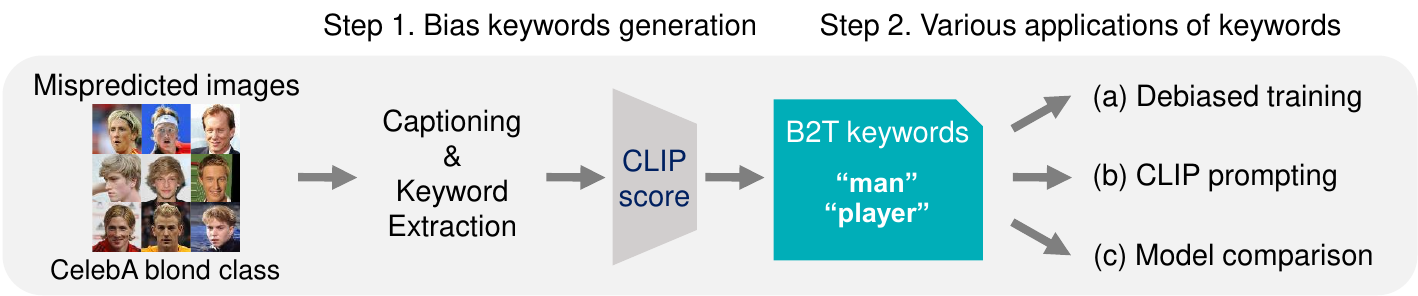}
% \vspace{-0.16in}
\caption{
\textbf{Method.}
(Step 1) B2T generates language descriptions from mispredicted images and extracts common keywords. We then verify whether these keywords indicate bias by measuring their similarity to the mispredicted images using a vision-language model like CLIP~\citep{radford2021learning}. (Step 2) The discovered keywords have various applications, including debiased training, CLIP prompting, and model comparison.
}\label{fig:method}
\vspace{-0.1in}
\end{figure*}

Previous research has attempted to identify visual biases by analyzing problematic samples~\citep{sohoni2020no, nam2020learning, liu2021just} or problematic attributes~\citep{singla2021understanding, singla2022salient, jain2022missingness}. However, these methods define biases indirectly, often relying on visualization or sample groups with specific statistics, and they require human supervision to express them in an explainable form.
To address this issue, recent research aimed at interpreting biases using vision-language models~\citep{radford2021learning}. Nonetheless, these studies have limitations in discovering and mitigating novel biases. Some studies~\citep{eyuboglu2022domino, jain2022distilling} retrieve the closest word from a predefined vocabulary, limiting their discovery to known biases. Others analyze neurons~\citep{hernandez2022natural} or images synthesized by generative models~\citep{wiles2022discovering} to comprehend biases. However, they focus on generating detailed captions explaining activated neurons or failure examples, which can help understand individual cases but hard to utilize for debiasing.

Instead, our main idea is to explain visual biases as \textit{keywords} by aggregating common traits from the language descriptions\footnote{
We used a pre-trained captioning model, but other language descriptions like hard prompt optimization~\citep{wen2023hard} can also be applied.
} of problematic images. Figure~\ref{fig:hook} illustrates our concept, with the keyword \keyword{flower} capturing the distinctive attributes of \keyword{ant} class images mispredicted as \keyword{bee.} This keyword form offers several advantages, providing a natural name for each bias group and easily incorporating with debiasing techniques using these group names.

\paragraph{Contribution.}
We introduce the \textit{Bias-to-Text} (B2T) framework, which identifies visual biases as keywords. To achieve this, we first generate language descriptions from mispredicted images and extract common keywords from these descriptions, suggesting potential biases. We then validate whether these keywords represent bias by measuring their similarity to the mispredicted images using a vision-language scoring model such as CLIP~\citep{radford2021learning}. By ensuring that the keywords align more closely with misclassified images rather than the correct ones, we can confirm they are biases.

\vspace{0.05in}
\noindent
We demonstrate that B2T can \textbf{discover biases} in image classifiers trained on various datasets (Section~\ref{sec:bias-discovery}):
\begin{itemize}
\item \textit{Known bias.}
B2T detects popular biases, such as gender bias in CelebA~\cite{liu2015deep}, background bias in Waterbirds~\cite{sagawa2020distributionally}, and distribution shifts in ImageNet-R~\cite{hendrycks2021many} and ImageNet-C~\cite{hendrycks2019benchmarking}. B2T keywords provide more fine-grained information for each bias, such as \keyword{bamboo,} representing the land background in Waterbirds. Moreover, these keywords can infer sample-wise bias labels, surpassing previous bias discovery approaches~\citep{eyuboglu2022domino,jain2022distilling}.
% Domino~\citep{eyuboglu2022domino}, and Failure Direction~\citep{jain2022distilling}
%
\item \textit{Novel bias.}
B2T uncovers novel biases in larger datasets, such as geographic bias in Dollar Street~\cite{rojas2022the} and contextual bias in ImageNet~\cite{deng2009imagenet}. 
For example, images with the keyword \keyword{flower} are predicted as \keyword{bee} instead of \keyword{ant} in ImageNet, indicating a contextual bias where bees are more commonly associated with flowers than ants.
% between bees and flower scenes.
% revealing a contextual bias where bees are more commonly associated with flowers than ants.
%
\end{itemize}

\vspace{0.05in}
\noindent
We then showcase that the bias keywords can be used for \textbf{various applications} (Section~\ref{sec:debias}):
\begin{itemize}
\item \textit{Debiased training.}
The keywords can be used to infer bias labels for each sample using the CLIP classifier. These labels are then used for debiased training, such as distributionally robust optimization (DRO)~\citep{sagawa2020distributionally}, and it outperforms previous debiasing approaches.
\item \textit{CLIP prompting.}
The keywords can be used to improve the CLIP zero-shot classifier. Prompting with fine-grained B2T keywords (e.g., \keyword{bamboo}) outperforms the previous strategy using group names (e.g., \keyword{land}).
\item \textit{Model comparison.}
The keywords can be used to compare the failure of different models. For example, ResNet~\cite{he2016deep} struggles more with complex scenes compared to ViT~\cite{dosovitskiy2020image}, as indicated by abstract keywords like \keyword{work out.}
\item \textit{Label diagnosis.}
B2T can detect issues in labels, such as mislabeling or label ambiguities. For example, we found that \keyword{bee} is often mislabeled as \keyword{fly} in ImageNet.
\end{itemize}

\vspace{0.05in}
\noindent
Lastly, we emphasize the robustness and versatility of our bias-to-text approach. First, we confirm that B2T keywords exhibit reasonable robustness across different captioning and similarity scoring models (Section~\ref{sec:ablation}), yet could be improved using advanced vision-language models like GPT-4~\citep{openai2023gpt}. Additionally, B2T can be extended beyond image classification, such as text-to-image generative models (Appendix~\ref{appx:gen}) and other computer vision tasks like object detection.

\section{Related Work}
\label{sec:related}

\textbf{Bias and fairness.}
Biases in datasets and models have long been issued in computer vision and machine learning~\citep{mehrabi2021survey}. Our goal is to study classifier failures for specific attributes or groups, known as spurious correlations~\citep{simon1954spurious}. These failures are closely related to fairness concerns, as models often perform poorly on particular gender~\citep{bolukbasi2016man, zhao2017men, hendricks2018women} or race~\citep{lee2018detecting, jalal2021fairness}. Such biases result from various sources, such as dataset imbalance~\citep{johnson2019survey, taori2022data}, or representational bias~\citep{wang2019balanced,zhao2023men,andrews2024ethical}, which is further exacerbated during model training. B2T aims to identify these fairness issues, providing \keyword{man} as a bias keyword for the \keyword{blond} class in CelebA~\citep{sagawa2020distributionally}.

Not only is bias related to fairness, but it also significantly impacts generalization, particularly in the presence of distribution shifts~\citep{muandet2013domain}. The ratio of majority to minority samples can vary, making the model susceptible to changes in their composition. This is closely connected to shortcut learning, where the model excessively relies on spurious features rather than core features~\cite{geirhos2020shortcut}. Various types of shortcuts exist, including texture bias~\cite{geirhos2019imagenet}, background bias~\cite{xiao2021noise}, and scene bias~\cite{mo2021object}. B2T could discover various types of shortcuts, as exemplified by \keyword{illustration} in ImageNet-R, \keyword{forest} in Waterbirds, \keyword{flower} for \keyword{ant} class in ImageNet.
\nocite{kang2022oamixer,moon2021masker}

\vspace{0.05in}
\noindent
\textbf{Bias discovery.}
Previous studies attempted to identify biases by analyzing problematic samples~\cite{d2022spotlight,krishnakumar2021udis,rajani2022seal,bao2022learning,atanov2022task,lee2022viscuit,jaipuria2022deeppic,wu2019errudite}.
Specifically, they detected biased samples by simply retrieving the mispredicted images~\cite{liu2021just} or utilizing embeddings or gradients~\cite{nam2020learning,sohoni2020no,agarwal2022estimating}.
% Specifically, they detected biased samples by simply retrieving the mispredicted images~\cite{liu2021just}, leveraging training statistics~\cite{nam2020learning}, clustering low-dimensional embeddings~\cite{sohoni2020no}, or using the variance of gradients~\cite{agarwal2022estimating}.
To uncover unknown biases, prior works iteratively trained a discoverer and classifier~\cite{li2022discover} or selected confident samples using two auxiliary biased models~\cite{zhao2022combating}. Another line of research analyzed problematic attributes to interpret spurious correlations and visualized them by highlighting specific regions~\citep{singla2021understanding, singla2022salient, jain2022missingness}, or generating traversal images alongside the attribute~\citep{li2021discover}. However, these methods still require human supervision to comprehend the common traits among failure cases, unlike B2T, which provides a practical keyword explanation.

\vspace{0.05in}
\noindent
\textbf{Bias discovery with language.}
Recent works describe bias using pre-trained vision-language models like CLIP~\citep{radford2021learning}. They define bias as an outlier (or slices) in the joint image-text embedding space~\citep{eyuboglu2022domino,jain2022distilling,zhang2023drml}. However, they match the outliers to a pre-defined bias vocabulary, limiting their ability to detect a single known bias. In contrast, B2T directly generates captions from images, potentially containing more detailed descriptions than the encoder embeddings. Thus, B2T effectively discovers multiple and fine-grained biases without the need for an iterative discovery procedure.

Other works analyze neurons~\citep{hernandez2022natural} or images synthesized by generative models~\citep{wiles2022discovering} to understand biases. In particular, \citet{wiles2022discovering} extract captions from synthesized images, similar to B2T. However, they provide detailed sentence descriptions, which are informative but not straightforward for debiasing. In contrast, the keyword explanation of B2T is more practical, as demonstrated in our applications, such as debiasing. Additionally, they need to specify a pair of true and mispredicted target classes, which may be challenging to scale if there are many classes. In contrast, B2T can find bias keywords for all failure cases simultaneously.

\vspace{0.05in}
\noindent
\textbf{Debiasing classifier.}
Numerous efforts have been made to mitigate biases of classifiers. DRO~\citep{rahimian2019distributionally, sagawa2020distributionally} is a popular approach that minimizes the loss over all bias groups. However, DRO requires bias annotations for all samples. Some works addressed this issue by inferring the bias group labels in an unsupervised manner~\cite{nam2020learning, liu2021just}. We demonstrate that the keyword explanations of B2T can infer bias labels using a zero-shot classifier like CLIP. This enables more accurate bias group estimation and improved debiased training compared to previous methods, as shown in Section~\ref{subsec:debias-dro}.

Moreover, we demonstrate a prompting strategy to debias CLIP using the B2T keywords, which are more fine-grained than those in prior work~\cite{zhang2022contrastive}, as shown in Section~\ref{subsec:debias-clip}.
\section{Bias-to-Text (B2T) Framework}
\label{sec:method}

In this section, we begin by defining the biases we aim to address~(\ref{sec:method-problem}).
Then, we introduce the Bias-to-Text (B2T) framework, which provides bias keywords using a captioning model and validates them with a scoring model (\ref{sec:method-overview}).
Finally, we validate the effect of the scoring model, showing that keywords with high scores tend to exhibit stronger bias (\ref{sec:method-score}).

\subsection{Problem formulation}
\label{sec:method-problem}

Image classifiers predict a class $y \in \mathcal{Y}$ for an image $x \in \mathcal{X}$. If images with attribute $a$ are frequently misclassified from their original class $y$, we refer to attribute $a$ as a bias associated with class $y$. Our goal is to identify this biased attribute $a$ in the \textit{keyword} explanation form.

The bias attributes include spurious correlation~\citep{simon1954spurious} or distribution shifts~\citep{shimodaira2000improving}. Spurious correlations lead models to rely on unintended decision rules (e.g., associating \keyword{blond} hair color with \keyword{man}), resulting in incorrect predictions when the rule does not apply~\citep{sagawa2020distributionally}. On the other hand, distribution shifts (e.g., style transfers like \keyword{illustration}) can impede model generalization in unseen samples~\citep{hendrycks2021many}.

% \begin{figure*}[t]
% \centering\small
% \includegraphics[width=\textwidth]{_cvpr2024/figures/fig_score.pdf}
% \caption{
% \textbf{Effect of the CLIP score.}
% We validate the effect of CLIP score on the waterbird class of the Waterbirds dataset.
% (a) CLIP score can identify incorrect bias keywords in image classifiers, showing low CLIP score near zero. 
% (b) The figures visualize the ROC curve of subgroup accuracy, defining the subgroup based on images with high CLIP similarity to specific keywords, while varying the similarity thresholds. The legend displays the B2T keywords alongside their corresponding CLIP scores in parentheses, with the AUROC of their respective curves denoted after the equal sign. Keywords with high CLIP scores tend to show low AUROC values for subgroup accuracy. 
% (c) The figures visualize the correlation between the CLIP score and AUROC of subgroup accuracy over B2T keywords, represented by colored dots. A significant negative correlation coefficient is observed between the two metrics
% }\label{fig:score}
% \end{figure*}

\begin{figure*}[t]
\centering\small
\begin{subfigure}{0.28\textwidth}
\centering\small
% \vspace{0.25in}
\includegraphics[width=\linewidth]{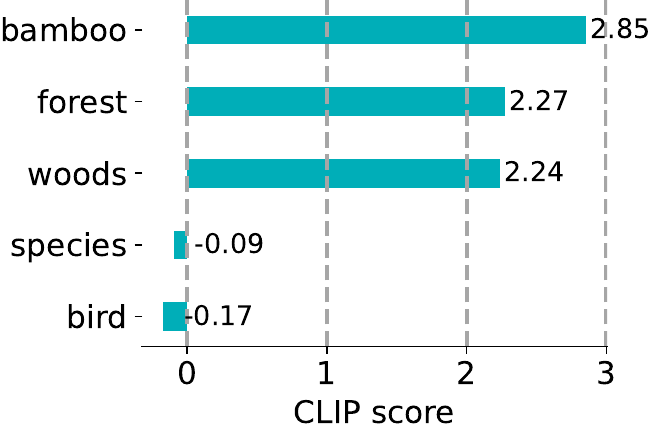}
\caption{CLIP score}
\end{subfigure}
\hspace{0.3in}
\begin{subfigure}{0.28\textwidth}
\centering\small
% \vspace{0.05in}
\includegraphics[width=\linewidth]{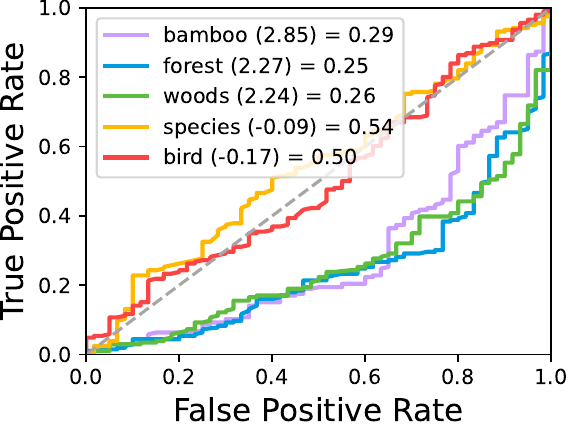}
\caption{ROC curve of subgroup accuracy}
\end{subfigure}
\hspace{0.3in}
\begin{subfigure}{0.28\textwidth}
\centering\small
% \vspace{0.05in}
\includegraphics[width=\linewidth]{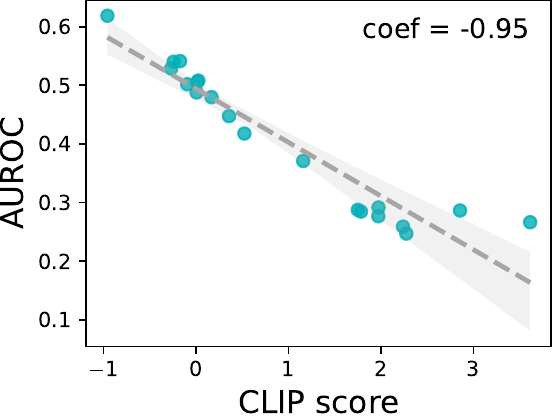}
\caption{Correlation of CLIP score and AUROC}
\end{subfigure}
\vspace{-0.05in}
\caption{
\textbf{Effect of the CLIP score (waterbird class).}
(a) The CLIP score can identify incorrect bias keywords, showing low CLIP scores near zero for non-bias keywords like \keyword{species.}
(b) The ROC curve represents subgroup accuracy, which defines the subgroup based on images with high CLIP similarity to specific keywords while varying the thresholds. The legend displays the B2T keywords alongside their corresponding CLIP scores in parentheses, with the AUROC of their respective curves denoted after the equal sign. Keywords with high CLIP scores tend to exhibit low subgroup accuracies, indicating they are biases. (c) Colored dots illustrate the negative correlation between the CLIP score and AUROC of subgroup accuracy over B2T keywords, indicating that a higher CLIP score implies stronger bias.
}\label{fig:score}
\vspace{-0.05in}
\end{figure*}

\subsection{Discovering bias keywords}
\label{sec:method-overview}

\textbf{Bias keywords.}
Our core idea is to extract keywords that represent biases. To achieve this, we extract common keywords from the language descriptions of class-wise \textit{mispredicted} images. Minority subgroups are those misclassified from the original class $y$ and thus often appear in these descriptions.
For example, in the case of the blonds vs. not-blonds classifier, the keyword \keyword{man} would frequently appear in the mispredicted images of the \keyword{blond} class.\footnote{
Technically, the keywords discovered by B2T are the \textit{opposite} of the biased concept. For example, B2T finds the keyword \keyword{man} for the blond class in a hair classifier. Here, \keyword{woman} is a bias-aligned (as an opposite of bias-conflicting) attribute following the definition of \citet{nam2020learning}.
} We employ a pre-trained captioning model~\cite{yu2022coca, li2023blip} to generate descriptions and extract common keywords.
We chose ClipCap~\cite{mokady2021clipcap} as our default captioning model because of its strong performance and fast inference speed (see Table~\ref{tab:ablation-captioning}), and apply the YAKE~\citep{campos2020yake} algorithm to extract keywords.

\vspace{0.05in}
\noindent
\textbf{CLIP score.}
We validate whether the keywords represent bias. To do this, we use a vision-language scoring model like CLIP~\citep{radford2021learning} that measures the similarities between keywords and the mispredicted images. The CLIP score ensures that keywords associated with a biased concept have a high CLIP score, while others have a low score. 
Specifically, we compare the CLIP embedding similarities between the keyword $a$ and images $x$ from $\mathcal{D}_\mathsf{wrong}$ and $\mathcal{D}_\mathsf{correct}$. These subsets of the class-wise validation set $\mathcal{D}$ correspond to the incorrect and correct predictions by the classifier, respectively.
Formally, the CLIP score is given by:
\begin{align}
s_\mathsf{CLIP}(a; \mathcal{D}) := \mathsf{sim}(a, \mathcal{D}_\mathsf{wrong}) - \mathsf{sim}(a, \mathcal{D}_\mathsf{correct}).
\label{eq:clip-score}
\end{align}
Here, $\mathsf{sim}(a, \mathcal{D})$ is the similarity between the keyword $a$ and the dataset $\mathcal{D}$, computed as the average cosine similarity between normalized embeddings of a word $f_\mathsf{text}(a)$ and images $f_\mathsf{image}(x)$ for $x \in \mathcal{D}$, where
\begin{align}
\mathsf{sim}(a, \mathcal{D}) := \frac{1}{|\mathcal{D}|} \sum_{x \in \mathcal{D}} f_\mathsf{image}(x) \, f_\mathsf{text}(a).
\end{align}
We referred to this as the CLIP score because we used CLIP as our default choice, but note that other vision-language models~\citep{cherti2022reproducible,li2022blip} also work well (see Table~\ref{tab:ablation-score}).

Further experimental details are provided in Appendix~\ref{appx:details}. While we primarily focus on image classifiers in this paper, our principle of interpreting visual biases as keywords can extend to other computer vision tasks, such as text-to-image generative models, as discussed in Appendix~\ref{appx:gen}.

\subsection{Validation of the CLIP score}
\label{sec:method-score}

We demonstrate the effect of the CLIP score in validating whether a keyword represents bias. Figure~\ref{fig:score} displays several analyses on the CLIP score using the waterbird class in the Waterbirds~\citep{sagawa2020distributionally} dataset. Panel (a) illustrates how the CLIP score identifies incorrect bias keywords. For instance, when the captioning model generates terms like \keyword{species} or \keyword{bird,} the CLIP score categorizes them as non-bias keywords due to their presence in both correctly and incorrectly predicted images, resulting in a low CLIP score.

Panel (b) displays the subgroup accuracy for each keyword. We use the CLIP similarity of individual samples associated with each keyword to define the subgroup. Subgroup accuracy is defined here as the AUROC calculated across different thresholds of CLIP similarity. Keywords with high CLIP scores (in parentheses) exhibit lower subgroup accuracies (after equal signs). For example, the keyword \keyword{bamboo} has a CLIP score of 2.85 and a subgroup accuracy of 0.29. In contrast, common keywords with CLIP scores near 0 (e.g., \keyword{bird}) demonstrate performance similar to random guessing (grey dotted line), suggesting that they are not biased.

Panel (c) visualizes the correlation between the CLIP score and subgroup accuracy (AUROC) for B2T keywords. These metrics have a high correlation coefficient of -0.95, indicating that the CLIP score reflects the severity of bias in keywords. See Appendix~\ref{appx:more} for further evaluations.

\begin{figure*}[t!]
\centering\small
\includegraphics[width=.95\textwidth]{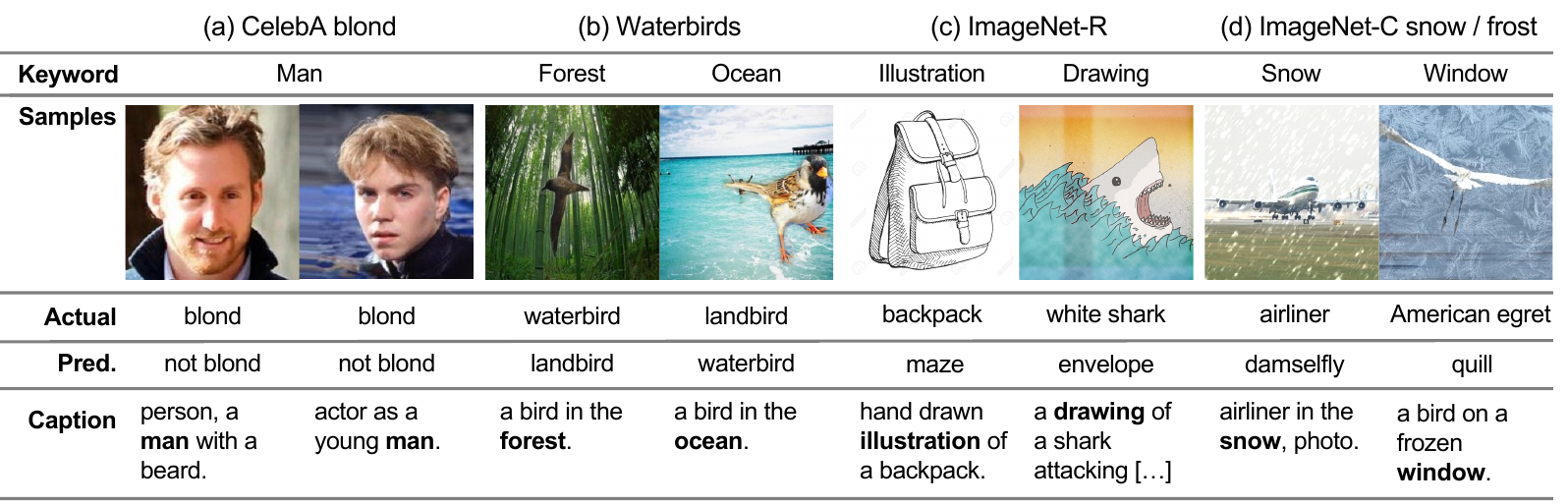}
\includegraphics[width=.95\textwidth]{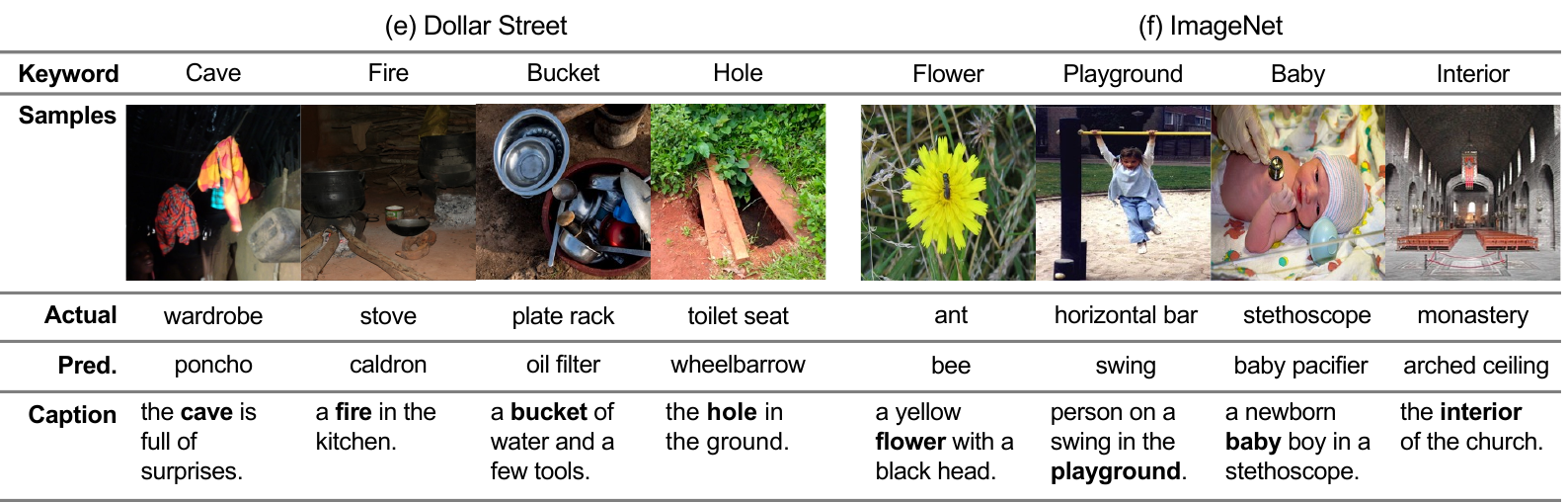}
\vspace{-0.08in}
% \vspace{-0.2in}
\caption{
\textbf{Discovered biases in image classifiers.}
Visual examples of mispredicted images, along with their corresponding bias keywords, captions, actual classes, and predicted classes. B2T successfully identified known biases, such as (a) gender bias in CelebA blond, (b) background bias in Waterbirds, and distribution shifts in (c) ImageNet-R with different styles, and (d) ImageNet-C with natural corruptions. B2T also uncovered novel biases in larger datasets, such as the spurious correlations between (e) the keyword \keyword{cave} and the wardrobe class, indicating geographical bias in Dollar Street, and (f) the keyword \keyword{flower} and the ant class, indicating contextual bias in ImageNet.
}\label{fig:cls-bias-known}
\vspace{-0.05in}
\end{figure*}

\section{Discovering Biases in Image Classifiers}
\label{sec:bias-discovery}

We demonstrate that B2T discovers visual biases in image classifiers trained on various datasets. First, we illustrate that B2T can identify known biases in benchmark datasets (\ref{subsec:known}). Then, we show how bias keywords infer sample-wise bias labels using the CLIP classifier, outperforming previous methods (\ref{subsec:known-eval}). Finally, we explore the capacity of B2T to uncover novel biases in larger datasets (\ref{subsec:unknown}).

\subsection{Can B2T identify the known biases?}
\label{subsec:known}

\textbf{Spurious correlation.}
We use B2T to analyze gender and background biases in the CelebA~\citep{liu2015deep} and Waterbirds~\citep{sagawa2020distributionally} datasets. CelebA contains facial images of celebrities with attribute annotations. Following \citet{sagawa2020distributionally}, we focus on classifying hair color as \keyword{blond} or \keyword{not blond.} Waterbirds comprise images of waterbirds and landbirds on land or water backgrounds. Here, we apply B2T to the empirical risk minimization (ERM) classifiers~\citep{sagawa2020distributionally}, which are known to be affected by spurious correlations.

Figure~\ref{fig:cls-bias-known} (a, b) displays the bias keywords.
B2T captures \keyword{man} for CelebA blond and \keyword{forest} and \keyword{ocean} for Waterbirds, revealing gender and background biases in each dataset.
Furthermore, B2T finds fine-grained keywords like \keyword{bamboo,} providing more detailed information than the original \keyword{land} background annotations.

\vspace{0.05in}
\noindent
\textbf{Distribution shifts.}
B2T can detect distribution shifts in ImageNet variants: ImageNet-R (rendition) \citep{hendrycks2021many}, which contains artistic images of ImageNet classes, and ImageNet-C (corruption) \citep{hendrycks2019benchmarking}, which contains noisy images of ImageNet classes. We use a supervised ResNet-50 \citep{he2016deep} classifier trained on ImageNet, which often struggles to generalize to these datasets, indicating its bias towards the training data. We apply B2T to the union of ImageNet and each variant (not class-wise) to identify the failures of the classifier.

Figure~\ref{fig:cls-bias-known} (c, d) displays the bias keywords.
For ImageNet-R, B2T captures keywords like \keyword{illustration} and \keyword{drawing,} with more detailed information such as \keyword{hand-drawn} and \keyword{vector art.}
For ImageNet-C, B2T captures keywords like \keyword{snow} for snow corruption, and \keyword{window} for frost corruption.
Here, the keyword \keyword{window} implies that the frozen images visually resemble those behind the window.

\subsection{Sample-wise bias labeling}
\label{subsec:known-eval}

We can infer sample-wise bias (or group) labels by applying the bias keywords to the CLIP zero-shot classifier. Specifically, we create prompts like \keyword{a photo of a [group],} where \keyword{[group]} represents the bias keywords, and we assign the label of each image to the nearest group.\footnote{
The prompt template depends on the dataset. For example, we use \keyword{a photo of a bird in the [group]} for Waterbirds. We select the bias keywords with CLIP scores larger than 1.0, like \keyword{forest} or \keyword{woods} for the waterbird class. See Table~\ref{tab:prompt_infer} in Appendix for the detailed prompt templates.
}

\begin{figure}[t]
\begin{subfigure}{0.23\textwidth}
\centering\small
\includegraphics[width=\linewidth]{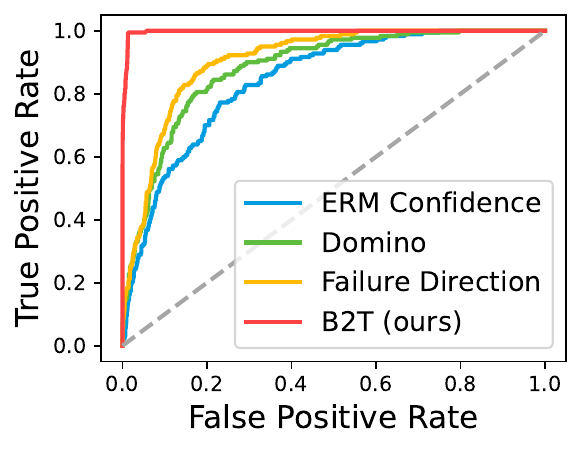}
\vspace{-0.18in}
\caption{CelebA blond}
\end{subfigure}~
\begin{subfigure}{0.23\textwidth}
\centering\small
\includegraphics[width=\linewidth]{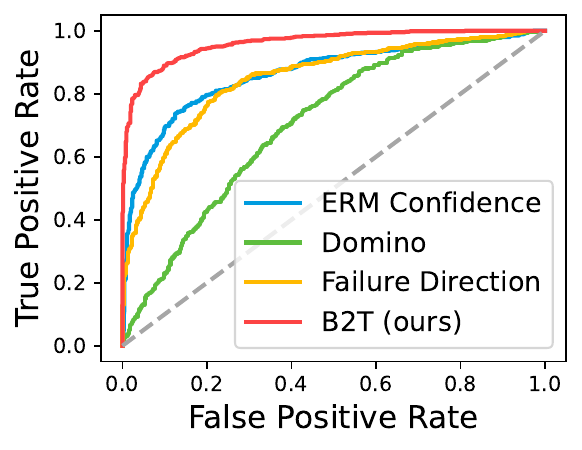}
\vspace{-0.18in}
\caption{Waterbird}
\end{subfigure}~
% \begin{subfigure}{0.24\textwidth}
% \centering\small
% \includegraphics[width=\linewidth]{figures/fig_cls-bias-auroc_3.pdf}
% \vspace{-0.18in}
% \caption{Landbird}
% \end{subfigure}
\vspace{-0.08in}
\caption{
\textbf{Comparison of bias discovery methods.}
The AUROC curves for (a) CelebA blond (male) and (b) Waterbirds (waterbirds on land), with parentheses indicating the corresponding minority groups. B2T outperforms prior works by a large margin.
}\label{fig:cls-bias-auroc}
\vspace{-0.2in}
\end{figure}

We then evaluate this sample-wise bias labeling in CelebA and Waterbirds, where ground-truth bias labels are available. We compare B2T with prior unsupervised bias discovery methods: JTT~\citep{liu2021just}, Domino~\citep{eyuboglu2022domino}, and Failure Direction~\citep{jain2022distilling}.
These methods use ERM confidence, GMMs, and SVMs to predict the bias labels, respectively.
Figure~\ref{fig:cls-bias-auroc} illustrates that B2T significantly outperforms prior methods, achieving near-optimal performance across all considered scenarios.

\subsection{Exploring novel biases in larger datasets} 
\label{subsec:unknown}

We apply B2T to discover novel biases in larger datasets. Note that B2T generates keywords from captioning models in a zero-shot manner, thus not requiring a pre-defined set of potential bias keywords, unlike prior works~\citep{eyuboglu2022domino,jain2022distilling}.

\vspace{0.05in}
\noindent
\textbf{Dollar Street.}
Dollar Street~\cite{rojas2022the} includes object images from countries with varying income levels. Previous studies have shown that classifiers perform poorly on objects from low-income countries~\cite{de2019does}. We aim to examine this geographic bias further by applying B2T to the validation set of Dollar Street using the ImageNet~\cite{deng2009imagenet} classifier. The classifier correctly predicted labels for objects from high-income countries but failed for low-income countries.

Figure~\ref{fig:cls-bias-known} (e) displays the bias keywords.
Here, B2T discovers bias keywords like \keyword{cave} for \keyword{wardrobe,} and \keyword{fire} for \keyword{stove} classes. Wardrobes from low-income countries are often in dark places resembling caves, and stoves from low-income countries often have a traditional design using fire. The keyword \keyword{bucket} for \keyword{plate rack} class suggests that buckets can be commonly used for stacking plates, and \keyword{hole} for \keyword{toilet seat} class suggests that the classifier is not familiar with squat toilets. These distinctions in objects across countries lead to geographical bias.

\vspace{0.05in}
\noindent
\textbf{ImageNet.}
We apply B2T to the ImageNet~\cite{deng2009imagenet} training set using the CLIP~\citep{radford2021learning} zero-shot classifier. We employ the 80-prompts ensemble strategy following the CLIP paper. We focus on investigating the highly confusing classes frequently misclassified as a specific class.

Figure~\ref{fig:cls-bias-known} (f) displays the bias keywords.
We discover contextual biases between objects in the scene. For instance, the classifier predicts \keyword{ant} with the keyword \keyword{flower} as \keyword{bee,} indicating a stronger association between flowers and bees than ants. The keyword \keyword{playground} implies that the classifier confuses a \keyword{horizontal bar} in the playground as a \keyword{swing.} The classifier confuses a \keyword{stethoscope} with the keyword \keyword{baby} as a \keyword{baby pacifier,} which is reasonable due to their similar appearances. In addition, we found the keywords \keyword{street} for the \keyword{plastic bag} class and \keyword{office} for the \keyword{notebook} class, suggesting that the classifier struggles in complex scenes with multiple objects.

\vspace{0.05in}
\noindent
\textbf{More examples.}
We provide additional visual examples in Appendix~\ref{appx:examples} and the lists of B2T keywords in Appendix~\ref{appx:table}.

\section{Applications of the B2T Keywords}
\label{sec:debias}

% We showcase that the keyword form of B2T offers various applications, including debiased training (~\ref{subsec:debias-dro}), CLIP prompting (~\ref{subsec:debias-dro}), model comparison (~\ref{subsec:debias-dro}), and label diagnosis (~\ref{subsec:debias-dro}).

We showcase that the keyword form of B2T offers various applications, including debiased DRO training, CLIP zero-shot prompting, model comparison, and label diagnosis.

\begin{table}[t]
\caption{
\textbf{Debiased DRO training.}
Worst-group and average accuracies (\%) of our debiased classifier (DRO-B2T) and prior works. GT denotes the usage of ground-truth bias labels for training, and bold denotes the best worst-group accuracy. B2T keywords enable accurate bias label prediction, facilitating effective DRO training.
}\label{tab:debias}
\vspace{-0.1in}
\centering\small
\newcolumntype{a}{>{\columncolor{mygreen! 10}}c}
\begin{tabular}{lc ac ac}
\toprule
& & \multicolumn{2}{c}{CelebA blond} & \multicolumn{2}{c}{Waterbirds} \\
\cmidrule(lr){3-4}\cmidrule(lr){5-6}
 Method & GT & Worst & Avg. & Worst & Avg. \\
\midrule
ERM                           & -   & 47.7\stdv{2.1} & 94.9  &  62.6\stdv{0.3} & 97.3 \\
LfF~\citep{nam2020learning}   & -   & 77.2\phantom{\stdv{x.x}} & 85.1  &  78.0\phantom{\stdv{x.x}} & 91.2 \\
GEORGE~\citep{sohoni2020no}   & -   & 54.9\stdv{1.9} & 94.6  &  76.2\stdv{2.0} & 95.7 \\
JTT~\citep{liu2021just}       & -   & 81.5\stdv{1.7} & 88.1  &  83.8\stdv{1.2} & 89.3 \\
CNC~\citep{zhang2022correct}  & -   & 88.8\stdv{0.9} & 89.9  &  88.5\stdv{0.3} & 90.9 \\
DRO-B2T (ours)            & -   & \textbf{90.4}\stdv{0.9} & 93.2  &  \textbf{90.7}\stdv{0.3} & 92.1 \\
\midrule
DRO~\citep{sagawa2020distributionally} & \cmark  & 90.0\stdv{1.5} & 93.3  &  89.9\stdv{1.3}  & 91.5 \\
\bottomrule
\end{tabular}
\end{table}

\subsection{Debiased DRO training}
\label{subsec:debias-dro}

Bias keywords can be used to train a debiased classifier. To be specific, we infer sample-wise bias labels as described in Section~\ref{subsec:known-eval} and apply them for training with DRO~\citep{sagawa2020distributionally}. We compare our DRO-B2T with various baselines, including ERM, DRO using the ground-truth (GT) bias labels, and debiased training methods that infer the group labels in an unsupervised manner: LfF~\citep{nam2020learning}, GEORGE~\citep{sohoni2020no}, JTT~\citep{liu2021just}, and CNC~\citep{zhang2022correct}. We excerpt values from the CNC paper.

% The values except DRO and DRO-B2T are excerpted from CNC paper.

Table~\ref{tab:debias} presents the worst-group and average accuracies. DRO-B2T outperforms the previous methods that infer group labels in an unsupervised manner, confirming the impact of B2T keywords. DRO-B2T also surpasses DRO using GT labels, possibly because of the noise in GT annotations. Check Appendix~\ref{appx:debias} for additional DRO experiments.

\begin{table}[t]
% \vspace{-0.05in}
\caption{
\textbf{CLIP zero-shot prompting.}
Worst-group and average accuracies (\%) of the CLIP zero-shot classifier using the base prompt or augmented ones: with the base group names (group) or B2T keywords with positive (B2T-pos) or negative (B2T-neg) CLIP scores. Bold indicates the best worst-group accuracy. B2T-pos improves worst-group accuracy, while B2T-neg harms. This implies that augmenting proper keywords to the prompts enhances the debiased accuracy of CLIP zero-shot inference.
}\label{tab:clip-IT-acc}
\vspace{-0.1in}
\centering\small
\newcolumntype{a}{>{\columncolor{mygreen! 10}}c}
\newcommand{\gap}[1]{{\scriptsize(#1)}}
% \resizebox{\linewidth}{!}{% 
\begin{tabular}{l ac ac}
\toprule
& \multicolumn{2}{c}{CelebA blond} & \multicolumn{2}{c}{Waterbirds} \\
 \cmidrule(lr){2-3}\cmidrule(lr){4-5}
& Worst & Avg. & Worst & Avg.  \\ 
\midrule
CLIP zero-shot & 76.2 & 85.2 & 50.3 & 72.7 \\
+ Group prompt~\citep{zhang2022contrastive} & 76.7 & 87.0 & 53.7 & 78.0 \\
\midrule
+ B2T-neg prompt & 72.9 & 88.0 & 45.4 & 70.8  \\
+ B2T-pos prompt (ours) & \textbf{80.0} & 87.2 & \textbf{61.7} & 76.9 \\
\bottomrule
\end{tabular}
\vspace{-0.05in}
\end{table}

\subsection{CLIP zero-shot prompting}\label{subsec:debias-clip}

Bias keywords can improve the CLIP zero-shot classifier by integrating them into prompts. In the original CLIP, the prompt template is \keyword{a photo of a [class].} We modify the prompt by adding a keyword, such as \keyword{a photo of a [class] in the [group],} where the keywords represent group names, as in the case of Waterbirds.\footnote{
See Table~\ref{tab:prompt_debias} in Appendix~\ref{appx:details} for the detailed prompt templates.
}
Here, we calculate the average text embeddings of prompts across all groups to get class embeddings and assign the image to the nearest class.

We augment the prompt with different sets of keywords to assess their importance. Specifically, we use B2T keywords with positive or negative CLIP scores, which we refer to as B2T-pos and B2T-neg, respectively. For instance, in the case of Waterbirds, we use \keyword{ocean} for B2T-pos and \keyword{bird} for B2T-neg.
B2T-pos keywords represent the minor subgroups, which would aid in recognizing them. We compare this approach with using the base group names, such as \keyword{water} background, as suggested in \citet{zhang2022contrastive}.

Table~\ref{tab:clip-IT-acc} presents the worst-group and average accuracies of the CLIP classifier. B2T-pos keywords enhance both worst-group and average accuracies. In contrast, base group names provide less assistance, and B2T-neg keywords even decrease the worst-group accuracy. This suggests that augmenting appropriate keywords to the prompts improves the debiased accuracy of CLIP zero-shot inference.

% , comparing the base prompt with augmented prompts containing keywords
% B2T-augmented prompts improve the fairness of ZS classifiers without fine-tuning.

\begin{figure}[t]
\centering\small
\includegraphics[width=\linewidth]{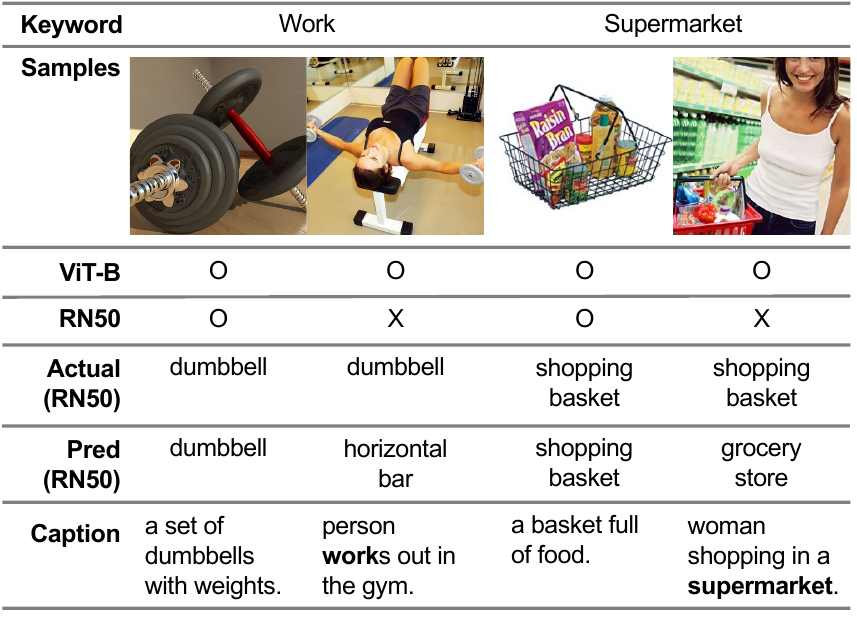}
\vspace{-0.25in}
\caption{
\textbf{Model comparison: ResNet vs. ViT.}
We compare the predictions made by ResNet and ViT, both trained and evaluated on ImageNet. We report their predicted labels and B2T keywords from ResNet. ViT excels at understanding global contexts and handling fine-grained classes than ResNet. For example, ResNet struggles with complex images whose B2T keywords represent abstract contexts like \keyword{work out} and \keyword{supermarket.}
}\label{fig:comp-model}
\vspace{-0.1in}
\end{figure}

\subsection{Model comparison}
Bias keywords can be used to analyze and compare different classifiers based on their keywords.

\vspace{0.05in}
\noindent
\textbf{Architecture: ResNet vs. ViT.} 
We compare ResNet~\cite{he2016deep} and ViT~\cite{dosovitskiy2020image} architectures. Recent studies claim that ViT is better than ResNet in understanding object shapes~\cite{naseer2021intriguing}. We further investigate this by examining bias keywords. We train and evaluate the models on ImageNet.

Figure~\ref{fig:comp-model} demonstrates the comparison results. ViT excels in understanding global contexts and fine-grained classes compared to ResNet. For instance, ViT successfully predicts complex images with abstract bias keywords like \keyword{work out.} We attribute this to the global self-attention of ViT, which allows for broader context consideration.

% \usepackage{booktabs}
% \usepackage{multirow}
% \usepackage{graphicx}
\begin{table}[t]
\centering\small
\caption{
% \textbf{Comparison of ERM and DRO training using B2T.}
\textbf{Model comparison: ERM vs. DRO.}
We compare biased (ERM) and debiased (DRO) classifiers on CelebA and Waterbirds. We present the CLIP scores for ERM, DRO, and the gap between them. We mark \xmark if the bias keyword is not found. In DRO, either the bias keyword is absent or its score is reduced; for example, the keyword \keyword{man} is no longer present in CelebA blond.
}\label{tab:training-comp}
\vspace{-0.1in}
% \resizebox{\linewidth}{!}{%
% \newcommand{\gap}[1]{{\color{red}#1}}
\newcommand{\gap}[1]{{#1}}
\newcolumntype{a}{>{\columncolor{mygreen! 10}}c}
% \begin{tabular}{@{}ccccc@{}}
\begin{tabular}{cccca}
\toprule
 & Keyword & ERM & DRO & Gap \\ \midrule
% \begin{tabular}[c]{@{}c@{}}CelebA\\      blond\end{tabular} & man & 1.06 & \xmark  & \gap{\xmark} \\
CelebA blond & man & 1.06 & \xmark  & \gap{\xmark} \\
\midrule
 & bamboo forest & 3.61 & \xmark  & \gap{\xmark} \\
 & bamboo & 2.85 & \xmark  & \gap{\xmark} \\
 & forest & 2.27 & 1.97 & \gap{-0.30} \\
\multirow{-4}{*}{Waterbird} & woods & 2.24 & 1.88 & \gap{-0.36} \\ \midrule
% \multirow{-5}{*}{Waterbird} & rainforest & 1.97 & \xmark  & \gap{\xmark} \\ \midrule
 & seagull & 3.10 & 1.85 & \gap{-1.24} \\
 & beach & 2.45 & 1.15 & \gap{-1.30} \\
 & water & 1.51 & 0.67 & \gap{-0.84} \\
\multirow{-4}{*}{Landbird} & lake & 1.25 & \xmark  & \gap{\xmark} \\ \midrule 
\end{tabular}%
% }
\vspace{-0.15in}
\end{table}

 %{-}1.22 & \xmark & \gap{\xmark}

  % \xmark & \gap{\xmark}

\vspace{0.05in}
\noindent
\textbf{Debiased training: ERM vs. DRO.} 
We compare biased and debiased training methods: ERM and DRO~\citep{sagawa2020distributionally}, on CelebA and Waterbirds. We list the bias keywords from ERM with CLIP scores higher than 1.0.

Table~\ref{tab:training-comp} illustrates the CLIP scores of ERM, DRO, and their gap. We mark \xmark if the keyword is not found. DRO indeed yields fewer bias keywords. For example, the keyword \keyword{man} is absent in the CelebA blond class, and the CLIP scores of highly biased keywords are reduced, such as from 3.10 to 1.85 for \keyword{seagull} in the landbird class.

\vspace{0.05in}
\noindent
\textbf{Additional model comparisons.} 
We present additional results in Appendix~\ref{appx:compare}, investigating the robustness of classifiers to distribution shifts. Our findings demonstrate that in multimodal learning, CLIP is more robust than ERM, and in self-supervised learning, MAE~\citep{he2022masked} exhibits better robustness, while DINO~\citep{caron2021emerging} shows similarity to ERM.

% \vspace{-0.1in}
\subsection{Label diagnosis}
\label{subsec:common-failure}
\begin{figure}[t]
\centering\small
\vspace{0.1in}
\includegraphics[width=\linewidth]{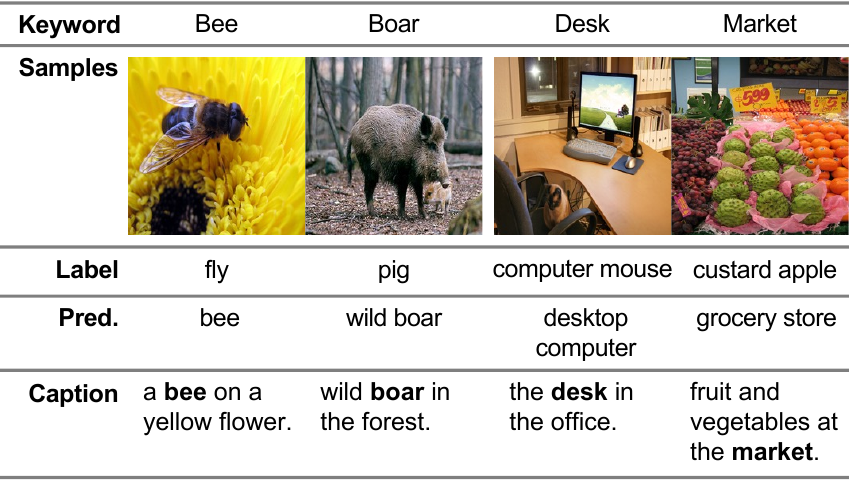}
\vspace{-0.25in}
\caption{
\textbf{Label diagnosis.}
% \todo{REVISED!}
We identify labeling errors, such as mislabeling and label ambiguities, in ImageNet using bias keywords. For example, the keyword \keyword{bee} implies that the images labeled as \keyword{fly} class are actually mislabeled. On the other hand, the keyword \keyword{desk} indicates that the images contain multiple objects, including both a \keyword{computer mouse} and a \keyword{desktop computer} on the desk, making it difficult to assign the appropriate class.
}\label{fig:common-failure}
\vspace{-0.15in}
\end{figure}

% \todo{REVISED!}
B2T can diagnose common labeling errors, such as mislabeling and label ambiguities. Previous studies have shown that ImageNet contains label errors~\citep{shankar20c}. We analyze these errors following the setup in Section~\ref{subsec:unknown}.

% using bias keywords.

% , following the setting in Section~\ref{subsec:unknown}.
% and illustrate class-specific instances under the same setting as in Section~\ref{subsec:unknown}.

Figure~\ref{fig:common-failure} visualizes examples. We found mislabeled images, such as \keyword{bee} and \keyword{boar} images labeled as \keyword{fly} and \keyword{pig,} respectively. We also found images with ambiguous labels, indicated by keywords \keyword{desk} and \keyword{market.} They are ambiguous as the scene contains multiple objects.

% the scene contains multiple objects, leading to ambiguity in labeling.

\section{Ablation Study}
\label{sec:ablation}

We study the effect of using different captioning and scoring models in our B2T framework. Details such as the architecture used for each model are stated in Appendix~\ref{appx:details}.

% Further details on ablation setups are stated in Appendix~\ref{appx:details}.

%
% We conduct ablation studies to analyze the impact of individual components in our proposed framework.

% \subsection{Ablation on captioning models}
\vspace{0.05in}
\noindent
\textbf{Captioning models.}
We study the robustness of the bias keywords across different captioning models: ClipCap~\cite{mokady2021clipcap}, BLIP~\cite{li2022blip}, BLIP-2~\cite{li2023blip}, CoCa~\cite{yu2022coca}, and LLaVA~\cite{liu2023visual}. 
% For LLaVA, we use the prompt \keyword{Describe the image in detail.} to extract captions.
% We study the robustness of the bias keywords across different captioning models. Specifically, we test ClipCap~\cite{mokady2021clipcap}
Table~\ref{tab:ablation-captioning} shows the results.
Different captioning models agree on severe biases while offering diverse fine-grained keywords. For example, all models capture major keywords like \keyword{man}, \keyword{forest}, and \keyword{beach}. Conversely, different models provide diverse fine-grained keywords, such as \keyword{rainforest} or \keyword{lake}; thus ensembling a few captioning models can diversify the discovered biased keywords. We use ClipCap as our default choice, given its strong performance and fast inference time. However, one could opt for advanced models like GPT-4~\citep{openai2023gpt} for improved captioning.

% ClipCap\footnote{\url{https://github.com/rmokady/CLIP_prefix_caption}}~\cite{mokady2021clipcap} model trained on Conceptual Captions~\cite{sharma2018conceptual} 

 % for CelebA blond class, Waterbirds waterbird class and landbird class, respectively.
 % For Waterbirds, different models yield divesre fine-grained keywords, thus ensembling a few captioning models together can diversify the discovered bias keywords. 
% Table~\ref{tab:ablation-captioning} compares B2T bias keywords across various captioning models:
% We leverage ClipCap unless otherwise mentioned, as it is the fastest model (4.31x faster than BLIP-2 model) while its captioning ability is comparable to other models.

% \subsection{Ablation on scoring models} 
\vspace{0.05in}
\noindent
\textbf{Scoring models.} 
We study the robustness of the CLIP score across different vision-language models. Specifically, we test CLIP trained on different datasets: OpenCLIP~\citep{cherti2022reproducible} trained on the LAION~\citep{schuhmann2022laion5b} dataset, and models with different architectures: BLIP~\citep{li2022blip} and BLIP-2~\citep{li2023blip}. 
% Specifically, we use CLIP ViT-L/14, OpenCLIP ViT-L/14 pretrained on LAION 2B, BLIP base, and BLIP-2 pretrain from LAVIS repository.~\footnote{
% \url{https://github.com/salesforce/BLIP}
% }
Table~\ref{tab:ablation-score} shows the results. The scoring models provide consistent rankings, with high scores for keywords like \keyword{man} or \keyword{bamboo forest.} We use CLIP as our default choice, but one could also consider the advanced models.

\vspace{0.05in}
\noindent
\textbf{Keyword extraction.} 
We use YAKE~\citep{campos2020yake} in our experiments, but other keyword extraction strategies, such as high-frequency words also perform well (see Appendix~\ref{appx:more}).
% or prompting GPT-3.5

% The keywords such as \keyword{man} for CelebA exhibit consistent rankings across all scoring models, while keywords unrelated to bias exhibit shifts in ranking.

% : CLIP~\citep{radford2021learning}, BLIP~\citep{li2022blip}, BLIP-2~\citep{li2023blip}, and OpenCLIP~\citep{cherti2022reproducible}.
% BLIP and BLIP-2 employ distinct training method from CLIP, highlighting the robustness of our proposed score under various VLP structures. In particular, BLIP mitigates noise in web image-text pairs through caption bootstrapping, and BLIP-2 leverages the capabilities of pre-trained large language models (LLMs). On the other hand, OpenCLIP is trained on different training dataset from CLIP, indicating that even when the primary source of bias (the training dataset) changes, our score consistently captures bias. Specifically, OpenCLIP is an open-source version of CLIP that is trained on the public LAION dataset~\citep{schuhmann2022laion5b}.
% \usepackage{booktabs}
% \usepackage{multirow}
% \usepackage{graphicx}
\begin{table}[t]
\caption{
\textbf{Ablation on different captioning models.}
B2T keywords discovered by different captioning models. We report the average inference time to extract a caption from a single image (in seconds on an RTX 3090 GPU) alongside the model names. The models consistently capture highly biased keywords such as \keyword{man,} \keyword{forest,} and \keyword{beach,} while different models may find diverse fine-grained keywords such as \keyword{rainforest} or \keyword{lake.}
% While all models capture \keyword{man}, \keyword{forest}, and \keyword{beach} for CelebA blond, Waterbirds waterbird and landbird, respectively, different models enable the diversity of fine-grained explanations of background biases. 
}\label{tab:ablation-captioning}
\vspace{-0.1in}
\centering\small
% \newcolumntype{a}{>{\columncolor{mygreen! 10}}c}
% \newcommand{\gap}[1]{{\scriptsize(#1)}}
\resizebox{\linewidth}{!}{%
% \begin{tabular}{@{}ccccccc@{}}
\begin{tabular}{ccccccc}
\toprule
 &  & ClipCap & BLIP & CoCa & BLIP-2 & LLaVA \\
\cmidrule(lr){3-3}\cmidrule(lr){4-4}\cmidrule(lr){5-5}\cmidrule(lr){6-6}\cmidrule(lr){7-7}
& Inference time& 0.13 sec & 0.20 sec & 0.34 sec & 0.56 sec & 1.90 sec \\ \midrule
% \begin{tabular}[c]{@{}c@{}}CelebA\\      blond\end{tabular} & man & O & O & O & O & O \\ \midrule
CelebA blond & man & O & O & O & O & O \\ \midrule
\multirow{4}{*}{Waterbird} 
 & forest & O & O & O & O & O \\
 & bamboo & O & O & O & O & O \\
 & woods & O & - & - & O & - \\
 & rainforest & O & - & - & - & - \\ \midrule
\multirow{4}{*}{Landbird}
 & beach & O & O & O & O & O \\
 & ocean & - & O & O & O & - \\
 & boat & - & O & O & O & O \\
 & lake & O & - & - & - & - \\ \bottomrule
\end{tabular}%
}
\vspace{-0.1in}
\end{table}
% \usepackage{graphicx}
% \usepackage{booktabs}
% \usepackage{multirow}
% \usepackage{graphicx}

\begin{table}[t]
\caption{
\textbf{Ablation on different scoring models.}
B2T keywords alongside their scores using different scoring models. The models provide consistent rankings, with high scores for keywords like \keyword{man} or \keyword{bamboo forest,} supporting their reliability.
% B2T keywords of CelebA blond with our proposed scores using different VLP models. All VLP models exhibit consistently high rankings for bias keywords \keyword{man} or \keyword{bamboo forest.}
}
\vspace{-0.1in}
\centering\small
\resizebox{\linewidth}{!}{%
% \begin{tabular}{@{}cccccc@{}}
\begin{tabular}{@{}cccccc@{}}
\toprule
                                &               & CLIP  & OpenCLIP & BLIP  & BLIP-2 \\ \midrule
\multirow{4}{*}{CelebA   blond} & man           & \phantom{-}1.06  & \phantom{-}2.23     & \phantom{-}1.19  & \phantom{-}4.04   \\
                                & player        & \phantom{-}0.35  & \phantom{-}1.30     & \phantom{-}0.74  & \phantom{-}2.67   \\
                                & face          & -0.28 & \phantom{-}0.44     & \phantom{-}0.49  & \phantom{-}1.46   \\
                                & actress       & -1.63 & -2.48    & -1.68 & -4.25  \\ \midrule
\multirow{4}{*}{Waterbird}      & bamboo forest & \phantom{-}3.61  & \phantom{-}4.68     & \phantom{-}5.22  & \phantom{-}9.85   \\
                                & woods         & \phantom{-}2.24  & \phantom{-}4.43     & \phantom{-}3.47  & \phantom{-}7.08   \\
                                & bird          & -0.09 & \phantom{-}0.67     & -0.03 & -0.70  \\
                                & pond          & -0.27 & -0.63    & -0.92 & -1.69  \\ \bottomrule
\end{tabular}%
}
\label{tab:ablation-score}
\vspace{-0.15in}
\end{table}

\section{Conclusion}

We propose B2T, a framework for identifying and mitigating biases through keyword explanation. The use of keywords offers several advantages, such as debiased training and model comparison. We hope that our B2T framework could assist in the responsible use of image recognition.
% We believe our concept of expressing visual biases as keywords would be a practical approach to addressing biases.

\vspace{0.05in}
\noindent
\textbf{Limitations.}
B2T relies on the recent advances in vision-language models, harnessing pre-trained captioning and scoring models. However, these models may not be perfect. For example, captioning models trained on web-crawled data may not generate informative descriptions in uncommon domains like medical and satellite images. Similarly, scoring models may not adequately capture image-text similarity due to limitations in their training data. Nevertheless, both models perform well in various scenarios, highlighting the practical merits of our B2T framework. Further discussions of limitations can be found in Appendix~\ref{appx:limitation}.

\vspace{0.05in}
\noindent
\textbf{Broader impacts.}
Bias and fairness research inherently have potential negative social impacts. We emphasize that B2T does not aim to \textit{fully automate} the discovery of biases but to \textit{assist} humans in making decisions based on the bias keywords. The final judgment is left to the users, who should also be monitored by a cross-verification system.

We illustrate sensitive examples - gender and geographic biases. It is crucial to note that our intention is to raise awareness and mitigate potential risks in real-world data.

\section*{Acknowledgements}
This work was supported by Institute of Information \& communications Technology Planning \& Evaluation (IITP) grant funded by the Korea government(MSIT) (No.2019-0-00075, Artificial Intelligence Graduate School Program(KAIST); No. 2022-0-00184, Development and Study of AI Technologies to Inexpensively Conform to Evolving Policy on Ethics). We thank Eunji Kim for constructive discussion.

{
    \small
    \bibliographystyle{ieeenat_fullname}
    \bibliography{main}
}

\clearpage

% \appendix
\begin{appendices}
\onecolumn

\begin{center}{\bf \Large Discovering and Mitigating Visual Biases through Keyword Explanation}\end{center}
\begin{center}{\Large Appendix}\end{center}

\hypersetup{linkcolor=black}
\etocdepthtag.toc{mtappendix}
\etocsettagdepth{mtchapter}{none}
\etocsettagdepth{mtappendix}{subsection}
\tableofcontents
\hypersetup{linkcolor=red}

\clearpage
\section{Implementation Details}
\label{appx:details}

% We use the ClipCap\footnote{\url{https://github.com/rmokady/CLIP_prefix_caption}}~\cite{mokady2021clipcap} model trained on Conceptual Captions~\cite{sharma2018conceptual} without beam search as our captioning model if not specified, and apply the YAKE\footnote{\url{https://github.com/LIAAD/yake}}~\cite{campos2020yake} algorithm to extract bias words from a corpus of mispredicted or generated samples. The maximum n-gram size is 3, and we select up to 20 keywords with a deduplication threshold of 0.9.

\textbf{Computation cost.}
With a single RTX 3090 GPU, it took approximately 30 minutes to generate captions for the CelebA validation set, which contains 19,867 images. Extracting keywords took 5 seconds, and deriving CLIP scores took 33 seconds.

\subsection{Bias discovery}
\subsubsection{Dataset details}
\textbf{CelebA blond.}
The CelebA~\cite{liu2015deep} dataset contains 19,867 validation images, and we use the ResNet-50~\citep{he2016deep} classifiers from the DRO repository.\footnote{\url{https://github.com/kohpangwei/group_DRO}} Specifically, we use the ERM and DRO models trained with a learning rate of 0.0001 and batch size of 128, achieving accuracies of 95.44\% and 90.40\% for the blond class, respectively.

\vspace{0.05in}
\noindent
\textbf{Waterbirds.}
The Waterbirds~\cite{sagawa2020distributionally} dataset contains 1,199 validation images, and we use the ResNet-50 classifiers from the DRO repository. Specifically, we use the ERM and DRO models trained with a learning rate of 0.001 and batch size of 128. ERM achieved accuracies of 86.66\% and 91.24\% for the waterbird and landbird classes, respectively.

\vspace{0.05in}
\noindent
\textbf{Dollar Street.}
The Dollar Street~\cite{rojas2022the} dataset contains a snapshot of the original web page on July 30th, 2019,\footnote{\url{https://github.com/greentfrapp/dollar-street-images}} and we use the ResNet-50 classifier trained on ImageNet for evaluation. We convert the class names of Dollar Street to ImageNet names using a mapping shown in Table~\ref{tab:match_dollar}.

\begin{table}[ht!]
\caption{
Conversion of class names from Dollar Street to ImageNet.
}\label{tab:match_dollar}
\vspace{-0.1in}
\centering
% \small
% \resizebox{\textwidth}{!}{% 
\begin{tabular}{ll}
\toprule
Dollar Street & ImageNet \\
\midrule
books & bookcase \\ 
computers & desktop computer \\ 
cups & tea cup \\ 
diapers & diaper \\ 
dish\_racks & plate rack \\ 
dishwashers & dishwasher \\ 
necklaces & necklace \\ 
stoves & stove \\ 
tables\_with\_food & dining table \\ 
toilet\_paper & toilet paper \\ 
toilets & toilet seat \\ 
wall\_clocks & wall clock \\ 
wardrobes & wardrobe \\ 
wheel\_barrows & barrow \\ 
wrist\_watches & digital watch \\ 
\bottomrule
\end{tabular}
\end{table}

\vspace{0.05in}
\noindent
\textbf{ImageNet and variants.}
ImageNet~\cite{deng2009imagenet} has 1,281,167 training images across 1,000 classes. We use CLIP zero-shot classifier with 80-prompts ensemble strategy and class names, following the CLIP paper. Specifically, with ResNet-50 architecture, the classifier achieves an accuracy of 60.56\% for vanialla ImageNet. We apply B2T to the most challenging classes, where the classifier exhibits the lowest accuracy.

ImageNet-R~\cite{hendrycks2021many} consists of 30,000 validation images, representing artistic renditions of ImageNet classes. We use the full set of 1,000 classes to infer classifiers, while ImageNet-R samples belong to a subset of 200 ImageNet classes. ImageNet-C~\cite{hendrycks2019benchmarking} contains corrupted versions of the ImageNet validation set, including snow and frost corruptions. Each corrupted dataset has 50,000 images, corresponding to vanilla ImageNet. To extract B2T keywords, we sample 10\% of each validation set and combine them with an equal number of samples from the original vanilla ImageNet.

We use the ResNet-50 classifier trained on vanilla ImageNet with the classic training recipe (V1) from the PyTorch model hub, which achieved 76.15\% accuracy for vanilla ImageNet. It achieves 52.8\%, 64.6\%, and 67.7\% accuracy for ImageNet-R, ImageNet-C snow, and ImageNet-C frost, respectively.

\subsubsection{Inferring bias labels}
\label{subsec:detail-label}

\textbf{Domino.}
Domino~\citep{eyuboglu2022domino} identifies underperforming subgroups, referred to as \keyword{slices}, by employing a Gaussian mixture model (GMM) in the CLIP embedding space. Following the parameters suggested in the paper, we use a log-likelihood weight of $10$ for $y$ and $\hat{y}$, and set the number of slices to $2$. We train slicing functions on the validation data for each class and then apply these learned slicing functions to the test data, resulting in soft slice assignments. The soft slice assignments are utilized to construct the AUROC curve.

% Domino~\citep{eyuboglu2022domino} identifies underperforming subgroups (``slices'') by using automated slice discovery methods (SDMs) in the CLIP embedding space. We followed the suggested parameters from the paper, with a y log likelihood weight of $10$, a y hat log likelihood weight of $10$, and the number of slices set to $2$. We trained slicing functions on the validation data for each class and then applied these learned slicing functions to the test data, resulting in soft slice assignments. The soft slice assignments were used to construct the AUROC curve.

\vspace{0.05in}
\noindent
\textbf{Failure Direction.}
Failure Direction~\citep{jain2022distilling} distills model failure modes using a linear support vector machine (SVM) to identify error patterns and represents them as directions within the CLIP feature space. We train class-wise SVMs on the validation data to obtain decision values for the test data, which are then used to construct the AUROC curve.

% Failure Direction~\citep{jain2023distilling} distills the failure modes of a model by using a linear support vector machine (SVM) to identify consistent error patterns and represents these failure modes as directions within the CLIP feature space. We trained class-wise SVMs using the validation data and obtained the decision values for the test data. The decision values were used to construct the AUROC curve.

\vspace{0.05in}
\noindent
\textbf{B2T (ours).}
For the CelebA dataset, we determine whether a training sample belongs to the \keyword{man} group or not. For the Waterbirds dataset, we determine the background of a training sample as either \keyword{land} or \keyword{water}. To effectively utilize the zero-shot classifier, we employ several techniques. Firstly, we use the general templates provided in the official CLIP ImageNet zero-shot classification\footnote{\url{https://github.com/openai/CLIP/blob/main/notebooks/Prompt_Engineering_for_ImageNet.ipynb}}. Secondly, we incorporate dataset-specific templates for improved information extraction. Lastly, we employ various B2T keywords as group names for classification. The prompts are generated in the format of \texttt{"[general template]+[dataset-specific template]+[group name],"} such as \keyword{a photo of a bird in a forest}. We use the CLIP ResNet-50 model. Table~\ref{tab:prompt_infer} presents the complete list of prompt templates and group names used.

\subsection{Debiasing classifiers}

\textbf{Debiased DRO training.}
We train DRO-B2T models following the protocol of~\cite{sagawa2020distributionally}. We utilize the SGD optimizer with a momentum of 0.9 to train ImageNet pre-trained ResNet-50 models on both datasets. For the CelebA dataset, we use a batch size of 64, a learning rate of 1e-5, a weight decay of 0.1, a group adjustment of 0, and train for 50 epochs. For the Waterbirds dataset, we use a batch size of 128 and train for 300 epochs. We sweep the hyperparameters {(learning rate, weight decay, group adjustment)} in the search space  
\(\{(1e-3, 1e-4, 0), (1e-4, 0.1, 0), (1e-5, 1.0, 0), (1e-5, 1.0, 1), (1e-5, 1.0, 2), (1e-5, 1.0, 3), (1e-5, 1.0, 4), (1e-5, 1.0, 5)\}\)
with validation worst-group accuracy. We report the average and worst-group test accuracies at the epoch with the best validation worst-group accuracy.

\vspace{0.05in}
\noindent
\textbf{CLIP zero-shot prompting.}
We augment prompt templates by adding B2T-inferred bias keywords to the end. Additionally, we utilize general templates provided for ImageNet zero-shot classification and dataset-specific templates to leverage the CLIP zero-shot classifier.
Table~\ref{tab:prompt_debias} presents the complete augmented templates with bias keywords that have positive CLIP scores. For example, a prompt for the landbird class in the Waterbirds dataset is \keyword{a photo of a landbird in the forest.} We generate ensembles of all possible prompt combinations while inferring the group labels. We use a pre-trained CLIP model with a ResNet-50 image encoder.

\begin{table*}[ht!]
\caption{
Prompt designs for inferring group labels.
}\label{tab:prompt_infer}
\vspace{-0.1in}
\centering\small
% \resizebox{\textwidth}{!}{% 
\begin{tabular}[t]{l | p{0.35\textwidth} | p{0.2\textwidth}}
\toprule
Dataset & Dataset-wise Template & Group Name \\
\midrule
\multirow{1}{*}{CelebA} &
\begin{itemize}[itemsep=1pt,topsep=0pt,leftmargin=10pt]
    \item \texttt{[group name]}
    \item \texttt{[group name] celebrity}
\end{itemize} &
\begin{enumerate}[itemsep=10pt,topsep=0pt,leftmargin=12pt]
    \item Male
    \begin{itemize}[itemsep=1pt,topsep=0pt,leftmargin=10pt]
        \item \texttt{man}
        \item \texttt{male}
    \end{itemize}
    \item Non-male
    \begin{itemize}[itemsep=1pt,topsep=0pt,leftmargin=10pt]
        \item Empty string \texttt{""}
    \end{itemize}
\end{enumerate} \\
\midrule
Waterbirds &
\begin{itemize}[itemsep=1pt,topsep=0pt,leftmargin=10pt]
    \item \texttt{[group name]}
    \item \texttt{bird on [group name]}
    \item \texttt{bird on a [group name]}
    \item \texttt{bird and a [group name]}
    \item \texttt{fowl on [group name]}
    \item \texttt{fowl on a [group name]}
    \item \texttt{fowl and a [group name]}
\end{itemize} &
\begin{enumerate}[itemsep=10pt,topsep=0pt,leftmargin=12pt]
    \item Land background
    \begin{itemize}[itemsep=1pt,topsep=0pt,leftmargin=10pt]
        \item \texttt{forest}
        \item \texttt{woods}
        \item \texttt{tree}
        \item \texttt{branch}
    \end{itemize}
    \item Water background
    \begin{itemize}[itemsep=1pt,topsep=0pt,leftmargin=10pt]
        \item \texttt{ocean}
        \item \texttt{beach}
        \item \texttt{surfer}
        \item \texttt{boat}
        \item \texttt{dock}
        \item \texttt{water}
        \item \texttt{lake}
    \end{itemize}
\end{enumerate} \\
\bottomrule
\end{tabular}
\end{table*}

\begin{table*}[ht!]
\caption{
Prompt designs for debiaisng zero-shot classifiers.
}\label{tab:prompt_debias}
\vspace{-0.1in}
\centering\small
% \resizebox{\textwidth}{!}{% 
\begin{tabular}[t]{l | p{0.35\textwidth} | p{0.4\textwidth}}
\toprule
Dataset & Dataset-wise Template & Class Name \\
\midrule
\multirow{1}{*}{CelebA} &
\begin{itemize}[itemsep=1pt,topsep=0pt,leftmargin=10pt]
    \item \texttt{[class name]}
    \item \texttt{[class name] man}
    \item \texttt{[class name] player}
    \item \texttt{[class name] person}
    \item \texttt{[class name] artist}
    \item \texttt{[class name] comedy}
    \item \texttt{[class name] film}
    \item \texttt{[class name] actor}
    \item \texttt{[class name] face}
\end{itemize} &
\begin{enumerate}[itemsep=10pt,topsep=0pt,leftmargin=12pt]
    \item Blond
    \begin{itemize}[itemsep=1pt,topsep=0pt,leftmargin=10pt]
        \item \texttt{blond hair}
        \item \texttt{celebrity of blond hair}
    \end{itemize}
    \item Non blond
    \begin{itemize}[itemsep=1pt,topsep=0pt,leftmargin=10pt]
        \item \texttt{non blond hair}
        \item \texttt{celebrity of non blond hair}
    \end{itemize}
\end{enumerate} \\
\midrule
Waterbirds &
\begin{itemize}[itemsep=1pt,topsep=0pt,leftmargin=10pt]
    \item \texttt{[class name]}
    \item \texttt{[class name] on the forest}
    \item \texttt{[class name] with woods}
    \item \texttt{[class name] on a tree}
    \item \texttt{[class name] on a branch}
    \item \texttt{[class name] in the forest}
    \item \texttt{[class name] on the tree}
    \item \texttt{[class name] on the ocean}
    \item \texttt{[class name] on a beach}
    \item \texttt{[class name] on the lake}
    \item \texttt{[class name] with a surfer}
    \item \texttt{[class name] on the water}
    \item \texttt{[class name] on a boat}
    \item \texttt{[class name] on the dock}
    \item \texttt{[class name] on the rocks}
    \item \texttt{[class name] in the sunset}
    \item \texttt{[class name] with a kite}
    \item \texttt{[class name] on the sky}
    \item \texttt{[class name] is on flight}
    \item \texttt{[class name] is on flies}
\end{itemize} &
\begin{enumerate}[itemsep=10pt,topsep=0pt,leftmargin=12pt]
    \item Landbird
    \begin{itemize}[itemsep=1pt,topsep=0pt,leftmargin=10pt]
        \item \texttt{landbird}
    \end{itemize}
    \item Waterbird
    \begin{itemize}[itemsep=1pt,topsep=0pt,leftmargin=10pt]
        \item \texttt{waterbird}
    \end{itemize}
\end{enumerate} \\
\bottomrule
\end{tabular}
\end{table*}

\subsection{Ablation studies}

\textbf{Captioning models.}
We use the ClipCap\footnote{\url{https://github.com/rmokady/CLIP_prefix_caption}}~\cite{mokady2021clipcap} model trained on Conceptual Captions~\cite{sharma2018conceptual} without beam search as our captioning model if not specified. We employ the BLIP~\cite{li2022blip} base captioning model trained on COCO and BLIP-2 utilizing the OPT-2.7b architecture from the LAVIS repository~\footnote{\url{https://github.com/salesforce/LAVIS}}. For CoCa~\cite{yu2022coca}, we use ViT-L-14 backbone pretrained on the LAION-2b dataset from the open CLIP repository~\footnote{\url{https://github.com/mlfoundations/open_clip}}, and for LLaVA~\cite{liu2023visual}, we use v1.5-13B that was trained in September 2023.

\vspace{0.05in}
\noindent
\textbf{Scoring models.} 
We use the CLIP model with the ViT-L/14 backbone from the CLIP repository~\footnote{\url{https://github.com/openai/CLIP}}. We employ OpenCLIP~\citep{cherti2022reproducible} with the ViT-L/14 backbone trained on the LAION-2b dataset~\citep{schuhmann2022laion5b}, and the base version of BLIP~\citep{li2022blip} and the pretrain version of BLIP-2~\citep{li2023blip} from the LAVIS repository \footnote{\url{https://github.com/salesforce/LAVIS}}. 

\vspace{0.05in}
\noindent
\textbf{Keyword extraction.} 
We apply the YAKE\footnote{\url{https://github.com/LIAAD/yake}}~\cite{campos2020yake} algorithm to extract bias keywords from a corpus of mispredicted or generated samples. The maximum n-gram size is 3, and we select up to 20 keywords with a deduplication threshold of 0.9. For high-frequency words, we lemmatize each word using WordNet~\cite{miller1995wordnet} to count words.

\clearpage
\section{Extension to Generative Models}
\label{appx:gen}

We extend the B2T framework to text-to-image (T2I) generative models~\citep{rombach2022highresolution}. Here, we define biases as spurious correlations between input conditions and generated attributes~\citep{nam2022breaking}, i.e., unintended attributes not explicitly specified through prompts.

\subsection{B2T for text-to-image (T2I) generative models}
T2I generative models produce an image $x \in \mathcal{X}$ from a given text description $y \in \mathcal{Y}$. Our goal is to identify a biased attribute $a \in \mathcal{A}$ that is spuriously correlated with the input prompts, i.e., the generated images $x$ contain the biased attribute $a$ even though it is not explicitly described in $y$. For example, a generative model may produce only female images when conditioned on blond, suggesting that the attribute \keyword{woman} is spuriously correlated with the prompt \keyword{blond.}

\vspace{0.05in}
\noindent
\textbf{Bias keywords.}
To identify the biased attribute $a$, we extract common keywords from the captions of the \textit{generated} images, rather than the mispredicted ones for classifiers. The keywords that appear in the generated images can be either the intended text $y$ or unintended bias $a$, and the user can infer the candidate set of biased keywords. In the case of the generative model conditioned on the prompt \keyword{blond,} the keywords \keyword{woman} (as well as \keyword{blond}) will frequently appear.

% However, this simple approach has a potential issue in that the captioning model may exhibit bias, as we discussed for the classifiers. It may misleadingly produce certain words frequently, even though they are not the majority keywords in the generated images. 
% Therefore, we additionally define a metric that verifies whether the generated images indeed reflect the unintended bias keywords. 

\vspace{0.05in}
\noindent
\textbf{SD score.}
To validate whether the keywords represent biases, we define a score analogous to the CLIP score. Our score relies on the underlying generative model being a T2I diffusion model~\citep{ho2020denoising}, but it could be extended to other generative models in principle. In our experiments, we use Stable Diffusion (SD)~\citep{rombach2022highresolution} and refer to our metric as the SD score.

The SD score measures the diffusion score between generated images and the original prompts or bias keywords, ensuring that only keywords that are already present in the generated images (and thus possibly associated with biased attributes) have a low SD score. To calculate this score, we compare the diffusion scores of generated images $x$ conditioned on the original prompt $y$ or bias keywords $a$.
Intuitively, the diffusion score for the conditions $y$ and $a$ will be similar if the generated image $x$ already reflects the bias keyword.
The SD score is given by:
\begin{align}
s_\mathsf{SD}(a; y) := \frac{1}{|\mathcal{D}_y|} \sum_{x \in \mathcal{D}_y}
|| \mathsf{score}(x; a) - \mathsf{score}(x; y) ||.
\label{eq:sd-score}
\end{align}
Here, $\mathcal{D}_y$ is the set of generated images conditioned on text $y$, $\mathsf{score}(x;y)$ is the diffusion score of an image $x$ conditioned on text $y$ (i.e., the gradient on the data space to update an image $x$ to follow the condition $y$), and $||\cdot||$ denotes the $\ell_2$-norm. The SD score uses the diffusion score of the generative model itself and is thus not affected by the bias in off-the-shelf captioning models. Additionally, the SD score can be interpreted as a classifier that uses the T2I diffusion model to compare the classification confidence of an image $x$ towards the classes $y$ and $a$, as explored in \citep{clark2023text,li2023your}.

\subsection{Experimental results} 
% Detailed discussions can be found in Appendix~\ref{appx:bias-gen}, and the complete list of keywords is included in Appendix~\ref{appx:more-table-gen}.

\begin{figure}[ht]
\centering\small
\includegraphics[width=0.48\textwidth]{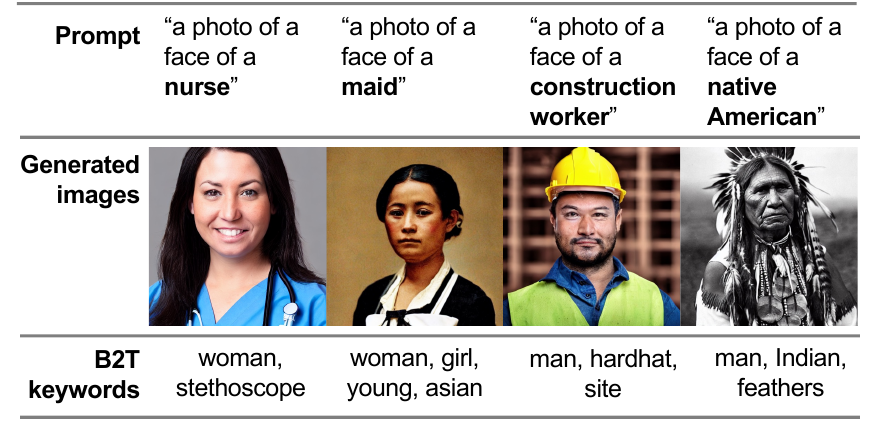}
\vspace{-0.1in}
\caption{
\textbf{Discovering biases in T2I generative models.} Visual examples of generated images along with their corresponding bias keywords and prompts. B2T successfully uncovers known biases, such as gender and race, that spuriously link to occupations~\citep{friedrich2023fair, luccioni2023stable}. B2T also discovers new spurious correlations, such as the pairings of \keyword{stethoscope} and \keyword{nurse,} suggesting that the model exhibits stereotypes based on the appearance of certain occupations or ethnicities.  
}\label{fig:gen-bias}
\end{figure}

\textbf{Bias discovery.}
We apply B2T on Stable Diffusion~\citep{rombach2022highresolution} using the prompts from~\citep{luccioni2023stable, friedrich2023fair}, resulting in the unfair generation of images. B2T recovers known biases, such as spurious correlations between occupations and gender or race. For instance, as shown in Figure~\ref{fig:gen-bias}, Stable Diffusion associates nurses with \keyword{women} and construction workers with \keyword{men,} indicating gender bias, and maids with \keyword{Asians,} indicating racial bias. Moreover, B2T uncovers unknown biases from the same prompts, such as the association of nurses with \keyword{stethoscope,} construction workers with \keyword{hat,} and Native Americans with \keyword{feathers,} suggesting that the model exhibits stereotypes based on the appearance of certain occupations and ethnicities.

\begin{figure}[ht]
\centering\small
\includegraphics[width=0.48\textwidth]{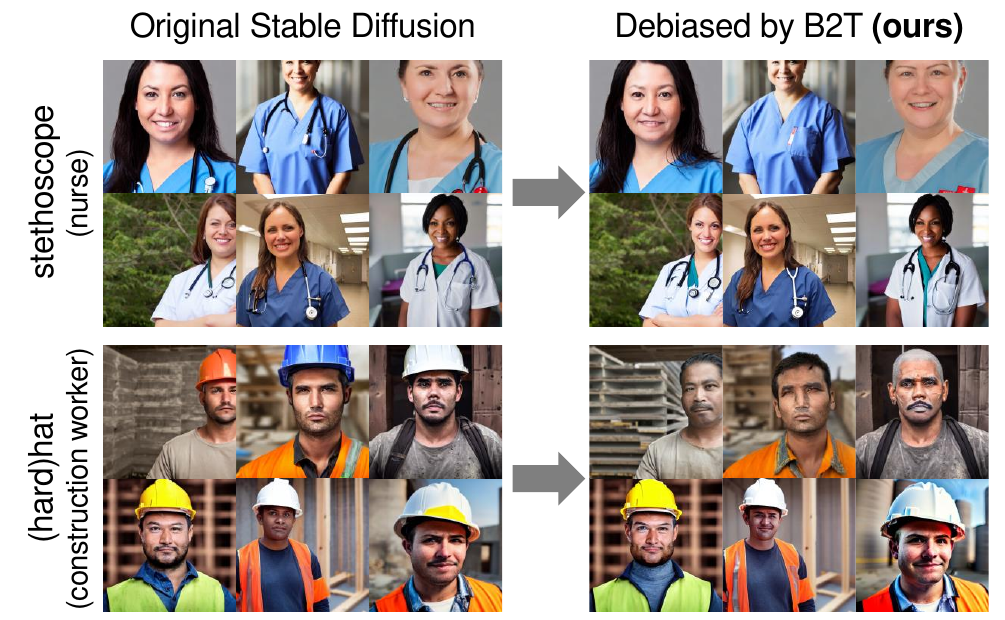}
\caption{
\textbf{Debiasing T2I diffusion models.}
We use the bias keywords discovered by B2T to debias the spurious correlations in Stable Diffusion. B2T effectively balances the generation of the unfair attribute \keyword{stethoscope} or \keyword{(hard)hat.}
}\label{fig:gen-debias}
\vspace{-0.05in}
\end{figure}

% \footnote{
% To be precise, we should add noise to the original image $x$ to follow the assumption of diffusion models. However, in practice, we found that clean images work well, as discussed further in Appendix~\ref{appx:ablation}.
% }
\vspace{0.05in}
\noindent
\textbf{Debiasing T2I diffusion.}
We use the bias keywords to debias a T2I diffusion model, Stable Diffusion~\citep{rombach2022highresolution}. To achieve this, we apply the Fair Diffusion~\citep{friedrich2023fair} algorithm, which adjusts the diffusion score that used to update images during generation, in order to regulate the effects of the specified keywords.
Figure~\ref{fig:gen-debias} demonstrates that Fair Diffusion, utilizing the bias keywords discovered by B2T, effectively eliminates the biases mentioned earlier. Our approach balances the unfair generation of images. We believe that B2T can facilitate the desirable use of fair T2I generative models.

\clearpage
\clearpage
\section{Additional DRO Results}
\label{appx:debias}

\subsection{Multi-class debiasing}

We conduct an additional experiment on datasets of more classes. We use the 2- and 10-class setups from the MetaShift~\citep{liang2022metashift} dataset, which aims to address spurious correlations between the cat and dog classes, associated with the indoor and outdoor attributes, respectively. First, we apply B2T to the ERM classifier and obtain outdoor keywords like \keyword{street} and \keyword{parked} for cats, as well as indoor keywords like \keyword{room} and \keyword{sleeping} for dogs. We then perform DRO training using these keywords and compare it with the baseline ERM and the oracle DRO using ground-truth labels. The table below displays the worst-group accuracies, with variations in the weights of minority subgroups (lower $p$ indicates stronger bias). DRO-B2T (ours) performs well for both the 2-class and 10-class scenarios.

\begin{table}[h]
\centering\small
\caption{\textbf{Multi-class debiasing.} DRO-B2T (ours) also works with multi-class debiasing scenarios.}
\vspace{-0.1in}
\begin{tabular}{lccccccc}
\toprule
   & &
  \multicolumn{3}{c}{\textbf{2 Class}} &
  \multicolumn{3}{c}{\textbf{10 Class}} \\ \cmidrule(l){3-5} \cmidrule(l){6-8}
               & \textbf{GT}   & $p$=12\% & $p$=6\% & $p$=1\% & $p$=12\% & $p$=6\% & $p$=1\% \\ \midrule
ERM            & -             & 50.00  & 47.92 & 37.50 & 68.58  & 67.01 & 63.19 \\
DRO-B2T (ours) & -             & 74.54	& 69.91	& 51.62	& 70.08  &	69.33 &	65.16 \\ \midrule
DRO            & \cmark        & 77.78	& 70.60	& 52.55	& 68.75  & 70.66 & 66.32 \\
\bottomrule
\end{tabular}
\end{table}

\subsection{Nonsensical groups}

Defining DRO subgroups by keywords does not pose a problem without human oversight: 1) keywords with high CLIP scores represent minorities, thus defining meaningful subgroups without supervision, and 2) even if the keywords are nonsensical, the subgroups become randomly sampled subsets, not affecting the outcome of DRO. To verify this, we perform an extra DRO experiment on the CelebA blond dataset, using a nonsensical keyword \keyword{face} alongside the bias keyword \keyword{man.} The table shows that the nonsensical keyword has no impact on the results. Lastly, human monitoring is still necessary due to the subjective nature of bias, and our goal is to assist rather than replace them.

\begin{table}[h]
\centering\small
\caption{\textbf{Nonsensical groups.} DRO-B2T (ours) also works with nonsensical group keywords.}
\vspace{-0.1in}
\begin{tabular}{ccc}
\toprule
\textbf{Keyword} & \textbf{Worst-group} & \textbf{Average} \\ \midrule
man             & 90.37\stdv{0.32}          & 93.02\stdv{0.31}      \\
man+face        & 90.00\stdv{0.96}          & 93.15\stdv{0.20}      \\ \bottomrule
\end{tabular}
\end{table}

% \setlength{\columnsep}{0.15in}
% \setlength{\intextsep}{0.05in}
% \begin{wraptable}{r}{0.5\linewidth}
% \resizebox{\linewidth}{!}{% 
% \begin{tabular}{@{}ccc@{}}
% \toprule
% % \textbf{keyword} & \textbf{Worst Group Acc.} & \textbf{Average Acc.} \\ \midrule
% \textbf{Keyword} & \textbf{Worst-group} & \textbf{Average} \\ \midrule
% man             & 90.37\stdv{0.32}          & 93.02\stdv{0.31}      \\
% man+face        & 90.00\stdv{0.96}          & 93.15\stdv{0.20}      \\ \bottomrule
% \end{tabular}%
% }\end{wraptable}%

\clearpage
\section{Additional Analyses}
\label{appx:more}

\subsection{Validation of the CLIP score}
\begin{figure}[ht]
\centering\small
\begin{subfigure}{0.28\textwidth}
\centering\small
% \vspace{0.25in}
\includegraphics[width=\linewidth]{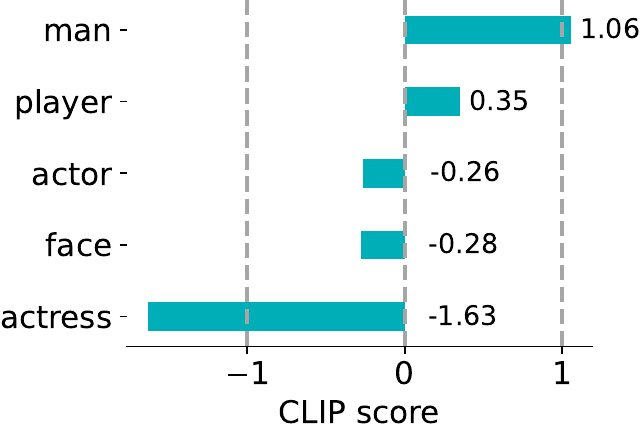}
\caption{CLIP score}
\end{subfigure}
\hspace{0.3in}
\begin{subfigure}{0.28\textwidth}
\centering\small
% \vspace{0.05in}
\includegraphics[width=\linewidth]{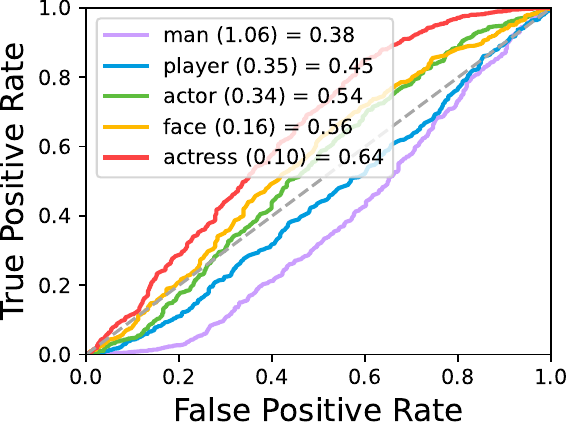}
\caption{ROC curve of subgroup accuracy}
\end{subfigure}
\hspace{0.3in}
\begin{subfigure}{0.28\textwidth}
\centering\small
% \vspace{0.05in}
\includegraphics[width=\linewidth]{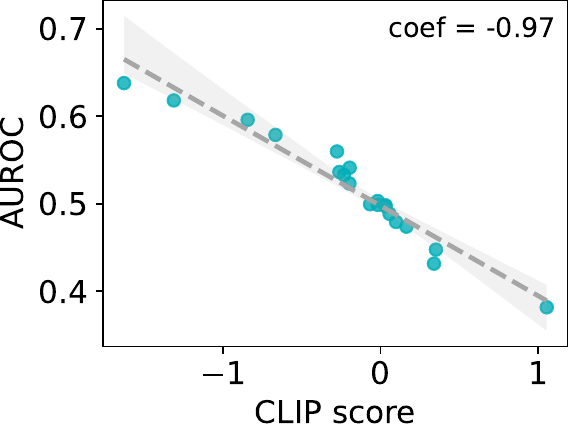}
\caption{Correlation of CLIP score and AUROC}
\end{subfigure}
\vspace{-0.05in}
\caption{
\textbf{Effect of the CLIP score (blond class in CelebA).} We can observe similar trends with the waterbird class.
% We demonstrate the effect of the CLIP score using the blond class of the CelebA dataset.
}\label{fig:score_blond}
\end{figure}

\begin{figure}[ht]
\centering\small
\begin{subfigure}{0.28\textwidth}
\centering\small
% \vspace{0.25in}
\includegraphics[width=\linewidth]{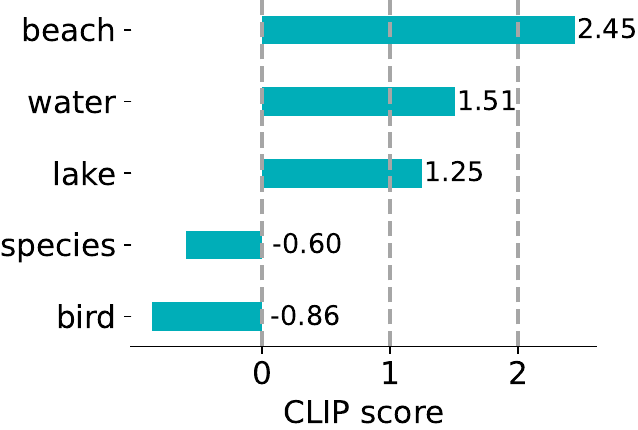}
\caption{CLIP score}
\end{subfigure}
\hspace{0.3in}
\begin{subfigure}{0.28\textwidth}
\centering\small
\includegraphics[width=\linewidth]{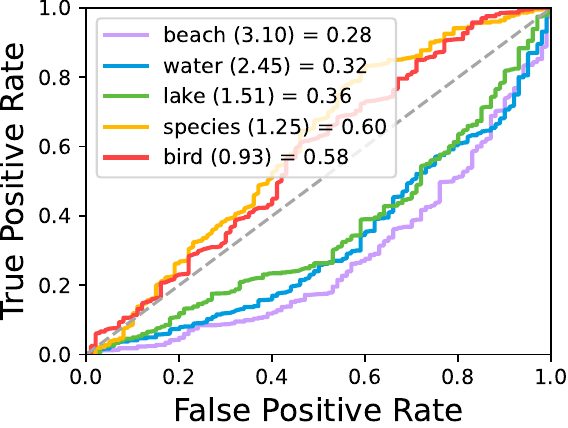}
\caption{ROC curve of subgroup accuracy}
\end{subfigure}
\hspace{0.3in}
\begin{subfigure}{0.28\textwidth}
\centering\small
\includegraphics[width=\linewidth]{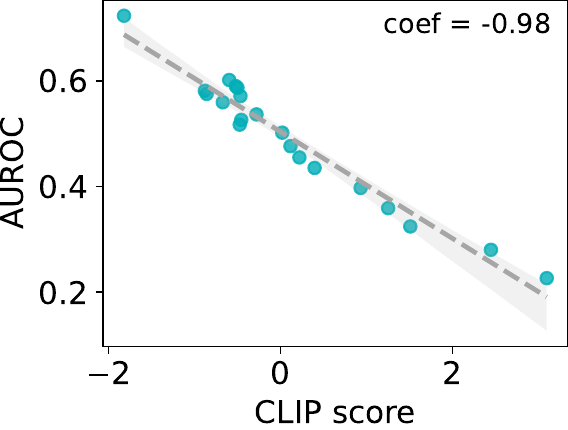}
\caption{Correlation of CLIP score and AUROC}
\end{subfigure}
\vspace{-0.05in}
\caption{
\textbf{Effect of the CLIP score (landbird class).} We can observe similar trends with the waterbird class.
}\label{fig:score_landbird}
\end{figure}

We demonstrate the effect of the CLIP score using the blond class of the CelebA dataset in figure~\ref{fig:score_blond} and the landbird class of the Waterbirds dataset in figure~\ref{fig:score_landbird}.

\subsection{Comparison of bias discovery methods}
\begin{figure}[ht]
\begin{subfigure}{0.32\textwidth}
\centering\small
\includegraphics[width=\linewidth]{figures/fig_cls-bias-auroc_1.pdf}
\vspace{-0.18in}
\caption{CelebA blond}
\end{subfigure}~
\begin{subfigure}{0.32\textwidth}
\centering\small
\includegraphics[width=\linewidth]{figures/fig_cls-bias-auroc_2.pdf}
\vspace{-0.18in}
\caption{Waterbird}
\end{subfigure}~
\begin{subfigure}{0.32\textwidth}
\centering\small
\includegraphics[width=\linewidth]{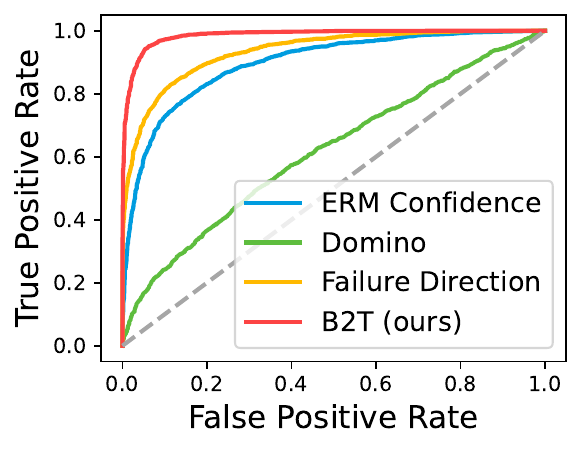}
\vspace{-0.18in}
\caption{Landbird}
\end{subfigure}
\caption{
\textbf{Comparison of bias discovery methods.}
The AUROC curves for (a) CelebA blond (male), (b) Waterbird (waterbird on land), and (c) Landbird (landbird on water) classes, with parentheses indicating the corresponding minority groups. B2T outperforms prior works by a large margin.
}\label{fig:additional-bias-auroc}
\vspace{-0.1in}
\end{figure}

We compare B2T with prior unsupervised bias discovery methods: JTT~\citep{liu2021just}, Domino~\citep{eyuboglu2022domino}, and Failure Direction~\citep{jain2022distilling}. Figure~\ref{fig:additional-bias-auroc} illustrates that B2T significantly outperforms prior methods, achieving near-optimal performance across all considered scenarios.

\subsection{Keyword extraction}
We compare the YAKE algorithm with a simple high-frequency and another popular keyword extraction algorithm, FRAKE. As depicted in Table~\ref{tab:additional_freq}, the extracted keywords are mostly shared across different methods. We observe significant biases such as \keyword{man} in CelebA blond or \keyword{forest} and \keyword{water} for Waterbirds waterbird and landbird class, respectively, across all keyword extraction methods. The 20 keywords for each method are reported in Table~\ref{tab:additional_freq}.

\begin{table*}[ht!]
\caption{Different keyword extraction methods}\label{tab:additional_freq}
\centering\small
\begin{subtable}{\textwidth}
\centering\small
\caption{High Frequency}
\vspace{-0.04in}
\begin{tabular}{@{}rl@{}}
\toprule
\multicolumn{1}{l}{\phantom{CelebA blond}} & \multicolumn{1}{c}{Keywords} \\ \midrule
CelebA blond &
  \begin{tabular}[c]{@{}l@{}}actor, person, hair, film, premiere, player, actress, face, \\ model, comedy, former, love, woman, like, artist, style, \\ man, want, first, contestant\end{tabular} \\\midrule
Waterbird &
  \begin{tabular}[c]{@{}l@{}}specie, biological, bird, tree, garden, person, forest, saw, \\ prey, one, wood, bamboo, wild, rainforest, paradise, pond, \\ rock, wall, selected, art\end{tabular} \\\midrule
Landbird &
  \begin{tabular}[c]{@{}l@{}}specie, biological, beach, bird, person, water, fly, \\ seagull, rock, sky, dog, seen, lake, city, pond, parrot, \\ yellow, one, saw, sunset\end{tabular} \\ \bottomrule
\end{tabular}%
\end{subtable}
\begin{subtable}{\textwidth}
\vspace{0.2in}
\centering\small
\caption{FRAKE}
\vspace{-0.04in}
\begin{tabular}{@{}rl@{}}
\toprule
\multicolumn{1}{l}{\phantom{CelebA blond}} & \multicolumn{1}{c}{Keywords} \\ \midrule
CelebA blond &
  \begin{tabular}[c]{@{}l@{}}actor person, actor premiere comedy film, person model \\ 
  actress, actor, person, want hair like, hair, player, film, \\ premiere, actress, model, face, comedy, love, \\ man, like, style, artist, contestant\end{tabular} \\ \midrule
Waterbird &
  \begin{tabular}[c]{@{}l@{}}biological species bird prey, biological bamboo forest, \\ species bamboo forest, biological, species, bird tree, bird, \\ tree, person, rainforest, saw, garden, forest, photo, wild,\\ bamboo, trees, pond, prey, woods\end{tabular} \\ \midrule
Landbird &
  \begin{tabular}[c]{@{}l@{}}biological species beach, bird flies water, bird beach, \\person beach, species, biological, bird, beach, person, \\water, flies, seagull, sits, sky, sunset, sea, paraglider, rocks, \\flight, city\end{tabular} \\ \bottomrule
\end{tabular}
\end{subtable}
\end{table*}

\clearpage
\section{Additional Model Comparisons}
\label{appx:compare}

\textbf{Multimodal learning: ERM vs. CLIP.}

Table~\ref{tab:comp_clip} presents a comparison of bias keywords obtained from ImageNet-R using ViT-B models trained by ERM and CLIP. ERM identifies distribution shifts like \keyword{illustration} and \keyword{drawing} as bias keywords, which have high CLIP scores. In contrast, CLIP identifies different bias keywords such as \keyword{dog} and exhibits low CLIP scores. This suggests that CLIP is less affected by distribution shifts compared to ERM.

\begin{table*}[ht!]
\caption{
Comparison of ERM vs. CLIP.
}\label{tab:comp_clip}
\vspace{-0.05in}
\centering\small
\begin{subtable}{0.44\textwidth}
\centering\small
\caption{ERM}\label{tab:compare_clip_erm}
% \resizebox{\textwidth}{!}{% 
\begin{tabular}{lc}
\toprule
% \phantom{white vector illustration}
& Score\\
\midrule
hand drawn illustration   & \phantom{-}2.02 \\
drawing                   & \phantom{-}1.61 \\
hand drawn                & \phantom{-}1.42 \\
vector illustration       & \phantom{-}1.38 \\
tattoo                    & \phantom{-}1.27 \\
white vector illustration & \phantom{-}1.22 \\
illustration              & \phantom{-}1.09 \\
sketch                    & \phantom{-}1.02 \\
step by step              & \phantom{-}0.53 \\
digital art               & \phantom{-}0.31 \\
\bottomrule
\end{tabular}
\end{subtable}
\begin{subtable}{0.44\textwidth}
\centering\small
\caption{CLIP}
% \vspace{-0.04in}
% \resizebox{\textwidth}{!}{% 
\begin{tabular}{lc}
\toprule
& Score \\
\midrule
dog                       & \phantom{-}0.64 \\
art                       & \phantom{-}0.55 \\
art selected              & \phantom{-}0.53 \\
person                    & \phantom{-}0.48 \\
tattoo                    & \phantom{-}0.48 \\
drawing                   & \phantom{-}0.45 \\
painting                  & \phantom{-}0.42 \\
step by step              & \phantom{-}0.36 \\
made                      & \phantom{-}0.31 \\
digital art selected      & \phantom{-}0.30 \\
\bottomrule
\end{tabular}
\end{subtable}
\end{table*}

\vspace{0.05in}
\noindent
\textbf{Self-supervised learning: ERM vs. DINO vs. MAE.}

Table~\ref{tab:comp_ssl} presents a comparison of bias keywords obtained from ImageNet-R using ViT-B models trained by DINO~\citep{caron2021emerging} and MAE~\citep{he2022masked}, along with ERM mentioned earlier. DINO provides similar bias keywords to ERM, while MAE provides keywords with low CLIP scores. Intuitively, both ERM and DINO demonstrate less robustness to distribution shifts than MAE.

\begin{table*}[ht!]
\caption{
Comparison of DINO vs. MAE.
}\label{tab:comp_ssl}
\vspace{-0.05in}
\centering\small
\begin{subtable}{0.44\textwidth}
\centering\small
\caption{DINO}
% \vspace{-0.04in}
% \resizebox{\textwidth}{!}{% 
\begin{tabular}{lc}
\toprule
\phantom{premiere of comedy}
& Score \\
\midrule
% vector art   illustration & 1.08 & 0.4 \\
% white vector illustration & 0.89 & 0.4 \\
% tattoo                    & 0.83 & 0.2 \\
% drawing                   & 0.76 & 0.5 \\
% illustration              & 0.71 & 0.3 \\
% vector illustration       & 0.59 & 0.4 \\
% art                       & 0.38 & 0.4 \\
% art selected              & 0.37 & 0.3 \\
% digital art selected      & 0.25 & 0.3 \\
% white background vector   & 0.23 & 0.1 \\
% hand drawn illustration   & \phantom{-}2.13 & \phantom{0}0.4 \\
% drawn vector illustration & \phantom{-}2.06 & \phantom{0}0.5 \\
% cartoon illustration      & \phantom{-}1.86 & \phantom{0}0.1 \\
% white vector illustration & \phantom{-}1.70 & \phantom{0}0.3 \\
% vector art illustration   & \phantom{-}1.63 & \phantom{0}0.5 \\
% vector illustration       & \phantom{-}1.53 & \phantom{0}0.3 \\
% tattoo                    & \phantom{-}1.48 & \phantom{0}0.2 \\
% white background vector   & \phantom{-}1.38 & \phantom{0}0.3 \\
% art                       & \phantom{-}0.97 & \phantom{0}0.3 \\
% digital art               & \phantom{-}0.90 & \phantom{0}0.3 \\
hand drawn illustration   & \phantom{-}2.13 \\
drawn vector illustration & \phantom{-}2.06 \\
cartoon illustration      & \phantom{-}1.86 \\
white vector illustration & \phantom{-}1.70 \\
vector art illustration   & \phantom{-}1.63 \\
vector illustration       & \phantom{-}1.53 \\
tattoo                    & \phantom{-}1.48 \\
white background vector   & \phantom{-}1.38 \\
art                       & \phantom{-}0.97 \\
digital art               & \phantom{-}0.90 \\
\bottomrule
\end{tabular}
\end{subtable}
\begin{subtable}{0.44\textwidth}
\centering\small
\caption{MAE}
% \vspace{-0.04in}
% \resizebox{\textwidth}{!}{% 
\begin{tabular}{lc}
\toprule
& Score \\
\midrule
% vector art illustration   & 0.90 & 0.6 \\
% tattoo                    & 0.88 & 0.2 \\
% drawing                   & 0.67 & 0.6 \\
% white vector illustration & 0.67 & 0.6 \\
% illustration              & 0.59 & 0.5 \\
% vector illustration       & 0.43 & 0.5 \\
% art                       & 0.39 & 0.5 \\
% art selected              & 0.31 & 0.5 \\
% digital art selected      & 0.17 & 0.5 \\
% white background vector   & 0.04 & 0.2 \\
% drawn vector illustration & \phantom{-}1.69 & \phantom{0}0.6 \\
% cartoon illustration      & \phantom{-}1.61 & \phantom{0}0.2 \\
% tattoo                    & \phantom{-}1.45 & \phantom{0}0.2 \\
% white vector illustration & \phantom{-}1.31 & \phantom{0}0.5 \\
% vector art illustration   & \phantom{-}1.27 & \phantom{0}0.6 \\
% vector illustration       & \phantom{-}1.20 & \phantom{0}0.5 \\
% drawing                   & \phantom{-}1.20 & \phantom{0}0.6 \\
% white background vector   & \phantom{-}1.05 & \phantom{0}0.4 \\
% art                       & \phantom{-}0.84 & \phantom{0}0.4 \\
% person                    & \phantom{-}0.78 & \phantom{0}0.7 \\
drawn vector illustration & \phantom{-}1.69 \\
cartoon illustration      & \phantom{-}1.61 \\
tattoo                    & \phantom{-}1.45 \\
white vector illustration & \phantom{-}1.31 \\
vector art illustration   & \phantom{-}1.27 \\
vector illustration       & \phantom{-}1.20 \\
drawing                   & \phantom{-}1.20 \\
white background vector   & \phantom{-}1.05 \\
art                       & \phantom{-}0.84 \\
person                    & \phantom{-}0.78 \\
\bottomrule
\end{tabular}
\end{subtable}
\end{table*}

\clearpage
\section{Further Discussion of Limitations}
\label{appx:limitation}

% \subsection{Limitations}

% \textbf{Limitations.}
We discover the bias of image classifiers using captioning (e.g., ClipCap \citep{mokady2021clipcap}) and scoring (e.g., CLIP \citep{radford2021learning}) models. However, there is a potential risk that these models themselves may be biased~\cite{agarwal2021evaluating,fang2022data}. Thus, users should not fully rely on the extracted captions, and the involvement of human juries remains essential in the development of fair machine learning systems.

For instance, ClipCap and CLIP are mostly trained on natural images, and are less effective for specialized domains~\citep{mo2023s} such as medical or satellite. To check this, we apply B2T to the ChestX-ray14~\cite{wang2017chestx} and FMoW~\citep{christie2018functional} datasets. We use classifiers publicly released in the ChexNet\footnote{
\url{https://github.com/arnoweng/CheXNet}
}~\cite{rajpurkar2017chexnet} and WILDS\footnote{
\url{https://worksheets.codalab.org/worksheets/0xa96b8749679944a5b4e4e7cf0ae61dc9}
}~\cite{koh2021wilds} codebases, utilizing the ERM classifier seed 0 for FMoW.

\begin{figure*}[ht!]
\centering\small
\includegraphics[width=.6\textwidth]{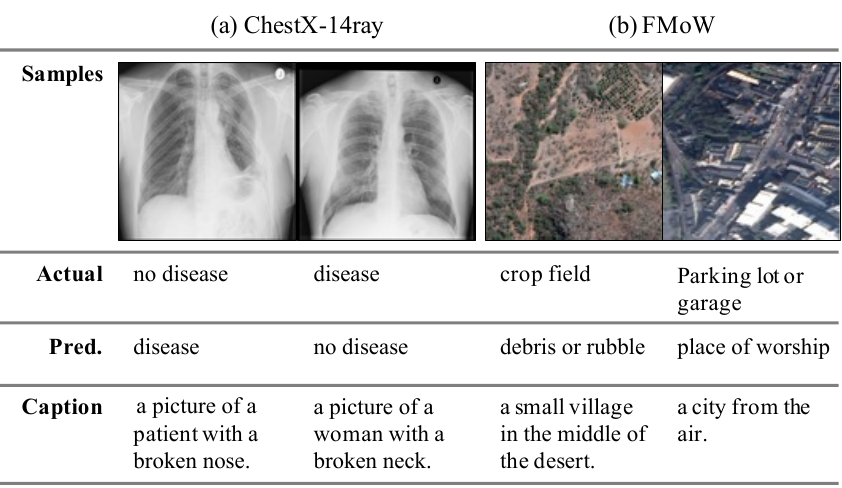}
\vspace{-0.1in}
\caption{
Visual examples of (a) ChestX-ray14 and (b) FMoW datasets.
}\label{fig:non-natural}
\end{figure*}

Figure~\ref{fig:non-natural} visualizes the images and their corresponding captions. ClipCap generates nonsensical captions, such as \keyword{broken nose} for chest images or trivial captions like \keyword{city from the air} for aerial-view images. Consequently, one must train a specialized captioning model to apply B2T effectively.

% \subsection{Broader impacts}

% Bias and fairness research inherently has potential negative social impacts. We emphasize that B2T does not aim to \textit{fully automate} the discovery of biases but to \textit{assist} humans in making decisions based on the bias keywords. The final judgment is left to the users, who should also be monitored by a cross-verification system.

% \textbf{Ethical concerns.}
% Bias and fairness research intrinsically has potential negative social impacts. We remark that B2T does not aim to \textit{fully automate} the procedure of discovering biases but \textit{assist} humans to make a decision based on the suggested candidates of bias keywords. We leave the final judgment to the users, who should also be monitored by a cross-verification system. We used some sensitive examples of gender and geographic biases. We emphasize that our purpose is to alert and prevent potential risks in real-world datasets, which are publicly accessible from Kaggle or Dollar Street websites.

% \textbf{Future works.}
% We discussed various applications of B2T, including discovering unknown biases in datasets, analyzing behaviors of classifiers, and debiasing zero-shot and full-shot classifiers. Applying B2T to more challenging real-world problems would be an exciting and impactful future direction. Also, one can extend B2T for creating a fairness or bias benchmark using the B2T bias keywords or designing a robust classifier by comparing models with B2T.

\clearpage
\section{Additional Visual Examples}
\label{appx:examples}
\vspace{-0.2in}

\begin{figure*}[ht!]
\centering\small
\includegraphics[width=\textwidth]{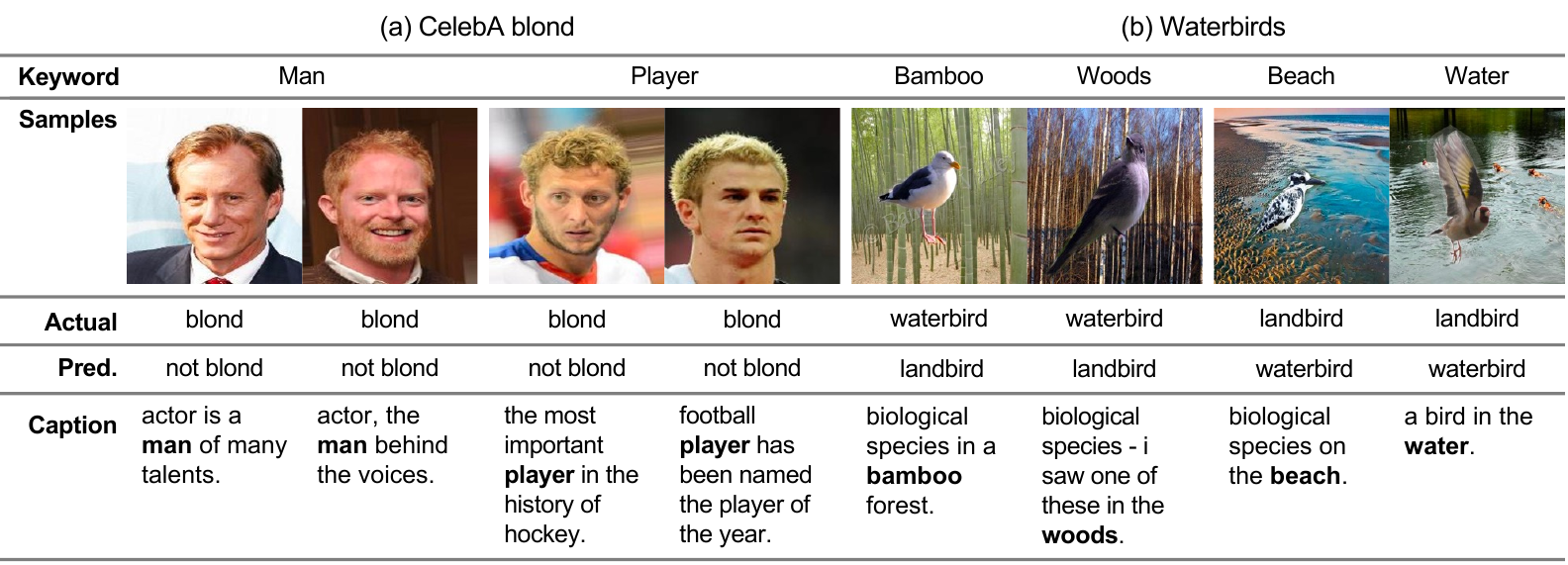}
\vspace{-0.2in}
\caption{
Additional visual examples of CelebA and Waterbirds.
}\label{fig:images-celeba}
\end{figure*}

% \begin{figure*}[ht!]
% \centering\small
% \includegraphics[width=\textwidth]{appendix/figures/more_celeba.pdf}
% \vspace{-0.2in}
% \caption{
% Additional visual examples of CelebA.
% }\label{fig:images-celeba}
% \end{figure*}

% \begin{figure*}[ht!]
% \centering\small
% \includegraphics[width=\textwidth]{appendix/figures/more_waterbirds.pdf}
% \vspace{-0.2in}
% \caption{
% Additional visual examples of Waterbirds.
% }\label{fig:images-waterbirds}
% \end{figure*}

\begin{figure*}[ht!]
\centering\small
\includegraphics[width=0.55\textwidth]{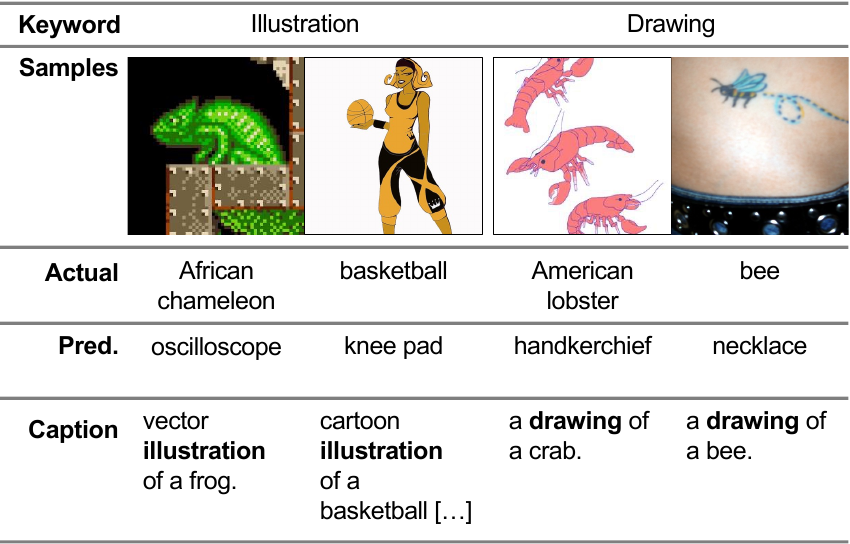}
\vspace{-0.1in}
\caption{
Additional visual examples of ImageNet-R.
}\label{fig:images-imagenet-variant}
\end{figure*}

\begin{figure*}[ht!]
\centering\small
\includegraphics[width=\textwidth]{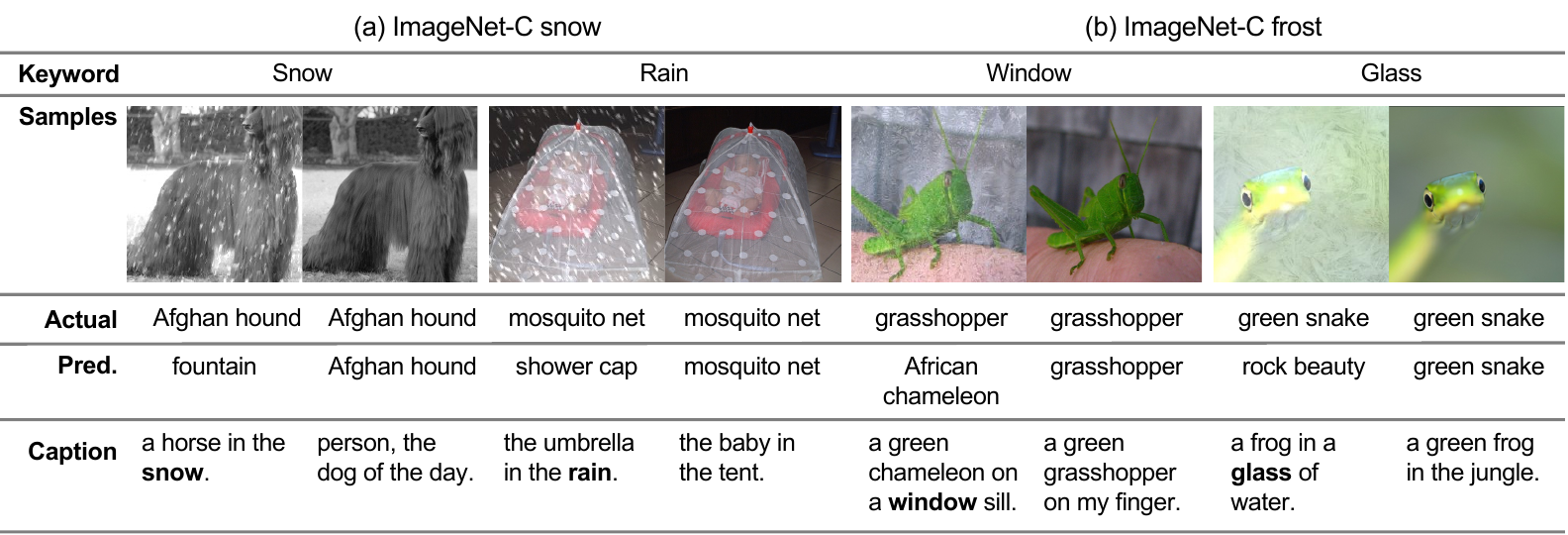}

\caption{
Additional visual examples of ImageNet-C.
}\label{fig:images-imagenet-variant}
\end{figure*}

\clearpage

\begin{figure*}[ht!]
\centering\small
\includegraphics[width=\textwidth]{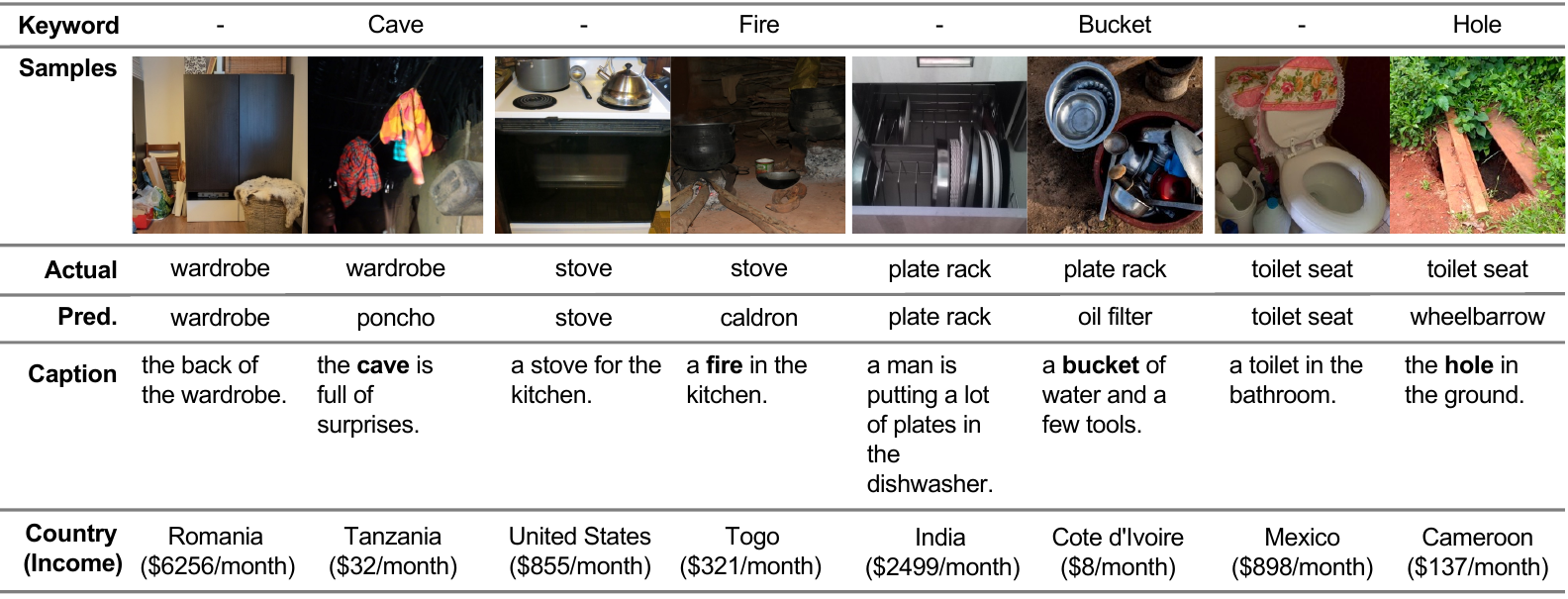}
\vspace{-0.2in}
\caption{
Visual examples of Dollar Street classes.
}\label{fig:images-dollar}
\end{figure*}

\begin{figure*}[ht!]
\centering\small
\includegraphics[width=\textwidth]{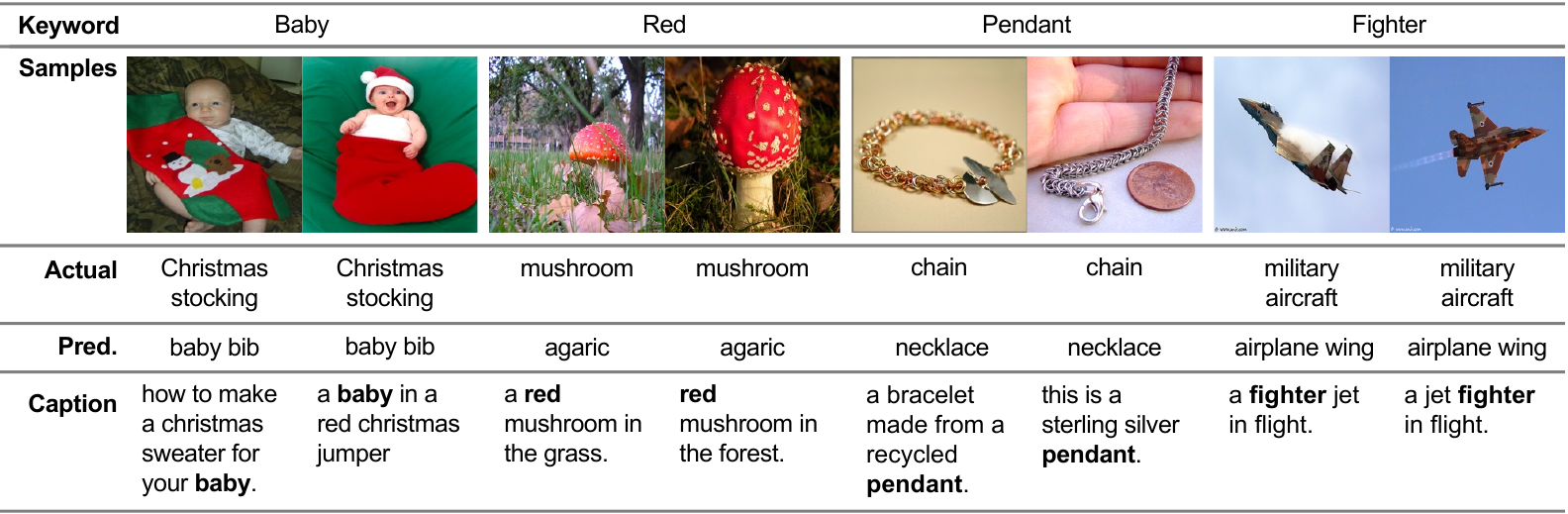}
~\\
\includegraphics[width=\textwidth]{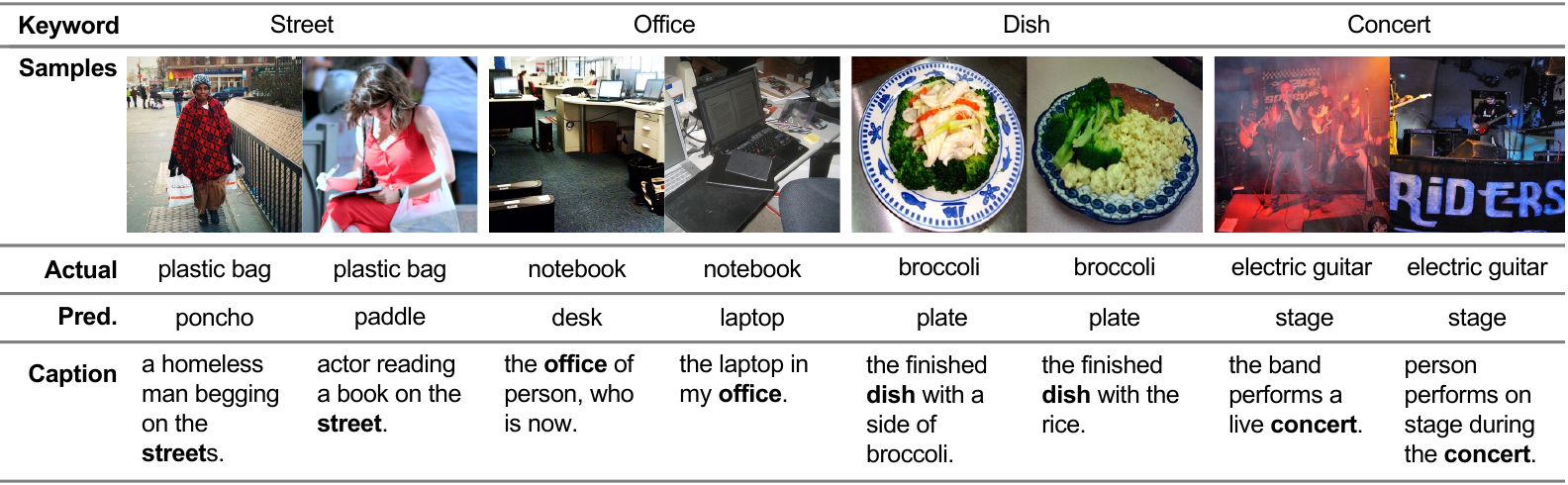}
\vspace{-0.2in}
\caption{
Additional visual examples of ImageNet classes.
}\label{fig:images-imagenet}
\end{figure*}

\clearpage
\section{Complete Lists of the B2T Keywords}
\label{appx:table}

Bias keywords from image classifiers, and their corresponding CLIP scores and subgroup accuracies. Higher CLIP scores and lower subgroup accuracies indicate more significant biases.

\begin{table*}[ht!]
\caption{Candidates of bias keywords for CelebA blond.}
\vspace{-0.05in}
\centering\small
\begin{subtable}{0.4\textwidth}
\centering\small
\caption{Blond (base acc.: 86.0)}
\vspace{-0.04in}
\begin{tabular}{lcc}
\toprule
\phantom{premiere of comedy}
& Score & Acc. \\
\midrule
man                 & \phantom{-}1.22 & 38.2 \\ 
player              & \phantom{-}0.42 & 27.8 \\
person              & \phantom{-}0.17 & 79.8 \\ 
artist              & \phantom{-}0.16 & 69.6 \\ 
comedy              & \phantom{-}0.16 & 88.2 \\ 
film                & \phantom{-}0.13 & 88.3 \\ 
actor               & \phantom{-}0.08 & 88.2 \\ 
face                & \phantom{-}0.06 & 88.5 \\ 
love                & \phantom{-}0.06 & 91.3 \\ 
clothing            & \phantom{-}0.05 & 93.5 \\ 
outfit              & \phantom{-}0.05 & 93.5 \\ 
hair                & \phantom{-}0.02 & 91.2 \\ 
style               & \phantom{-}0.00 & 92.2 \\ 
weight              & -0.06 & 93.6 \\ 
clothing style      & -0.08 & 93.5 \\ 
model               & -0.19 & 95.5 \\ 
premiere            & -0.52 & 89.1 \\ 
premiere of comedy  & -0.63 & 86.2 \\ 
model and actress   & -1.00 & 82.7 \\
actress             & -1.28 & 83.3 \\ 
\bottomrule
\end{tabular}
\end{subtable}
\begin{subtable}{0.4\textwidth}
\centering\small
\caption{Not blond (base acc.: 97.2)}
\vspace{-0.04in}
% \resizebox{\textwidth}{!}{% 
\begin{tabular}{lcc}
\toprule
\phantom{premiere of comedy}
& Score & Acc.\\
\midrule
model           & \phantom{-}0.50 & 96.9 \\
favorite outfit & \phantom{-}0.34 & 94.8 \\
hair            & \phantom{-}0.33 & 94.4 \\
love            & \phantom{-}0.17 & 96.7 \\
style           & \phantom{-}0.14 & 94.7 \\
premiere        & \phantom{-}0.11 & 98.0 \\
clothing style  & \phantom{-}0.09 & 94.8 \\
outfit          & \phantom{-}0.08 & 94.8 \\
favorite        & \phantom{-}0.08 & 94.8 \\
feet size       & \phantom{-}0.06 & 94.8 \\
clothing        & \phantom{-}0.06 & 94.8 \\
film            & \phantom{-}0.00 & 98.3 \\
weight          & -0.03 & 94.8 \\
face            & -0.05 & 97.3 \\
feet            & -0.06 & 94.8 \\
size            & -0.08 & 94.8 \\
comedy          & -0.25 & 96.5 \\
person          & -0.28 & 97.5 \\
bob             & -0.50 & 93.2 \\
actor           & -0.98 & 97.5 \\
\bottomrule
\end{tabular}
\end{subtable}
\end{table*}

\begin{table*}[ht!]
\caption{Candidates of bias keywords for Waterbirds.}
\vspace{-0.05in}
\centering\small
\begin{subtable}{0.4\textwidth}
\centering\small
\caption{Waterbird (base acc.: 75.6)}
\vspace{-0.04in}
\begin{tabular}{lcc}
\toprule
\phantom{premiere of comedy}
& Score & Acc. \\
\midrule
forest              & \phantom{-}2.12 & 61.5 \\ 
woods               & \phantom{-}1.94 & 62.5 \\ 
tree                & \phantom{-}1.45 & 41.7 \\ 
branch              & \phantom{-}1.20 & 35.7 \\ 
prey                & \phantom{-}0.20 & 70.0 \\ 
wild                & \phantom{-}0.19 & 75.0 \\ 
bird of prey        & -0.03 & 66.7 \\ 
species             & -0.05 & 74.2 \\ 
area                & -0.09 & \phantom{0}0.0 \\ 
biological species  & -0.11 & 74.2 \\ 
bird in flight      & -0.27 & 50.0 \\ 
biological          & -0.28 & 74.2 \\ 
bird                & -0.36 & 62.5 \\ 
person              & -0.41 & 81.3 \\ 
bird flying         & -0.42 & 75.0 \\ 
eagle               & -0.69 & 95.5 \\ 
bald                & -0.69 & 60.0 \\ 
snow                & -0.80 & 66.7 \\ 
great bird          & -0.80 & \phantom{0}0.0 \\ 
large bird          & -1.05 & 50.0 \\ 
\bottomrule
\end{tabular}
\end{subtable}
\begin{subtable}{0.4\textwidth}
\centering\small
\caption{Landbird (base acc.: 89.9)}
\vspace{-0.04in}
\begin{tabular}{lcc}
\toprule
\phantom{premiere of comedy}
& Score & Acc. \\
\midrule
ocean               & \phantom{-}3.41 & 44.4 \\
beach               & \phantom{-}2.83 & 74.7 \\
surfer              & \phantom{-}2.73 & 55.6 \\
boat                & \phantom{-}2.16 & 64.7 \\
dock                & \phantom{-}1.56 & 75.0 \\
water               & \phantom{-}1.38 & 75.0 \\
lake                & \phantom{-}1.17 & 80.0 \\
rocks               & \phantom{-}1.02 & 76.5 \\
sunset              & \phantom{-}0.88 & 70.0 \\
kite                & \phantom{-}0.67 & 64.6 \\ 
sky                 & \phantom{-}0.28 & 84.2 \\
flight              & \phantom{-}0.23 & 62.5 \\
flies               & -0.17 & 73.3 \\
person              & -0.38 & 86.9 \\ 
pond                & -0.47 & 87.0 \\ 
biological species  & -0.48 & 95.5 \\ 
biological          & -0.55 & 93.4 \\ 
species in flight   & -0.92 & 44.4 \\ 
species             & -0.97 & 93.4 \\ 
bird                & -1.64 & 93.8 \\ 
\bottomrule
\end{tabular}
\end{subtable}
\end{table*}

\begin{table*}[ht!]
\caption{Candidates of bias keywords for ImageNet-R and ImageNet-C.}
\vspace{-0.05in}
\centering\small
\begin{subtable}{.36\textwidth}
\centering\small
\caption{ImageNet-R (base acc.: 52.8)}
\vspace{-0.04in}
% \resizebox{\textwidth}{!}{% 
\begin{tabular}{lcc}
\toprule
\phantom{white background vector}
& Score & Acc. \\
\midrule
hand drawn illustration   & \phantom{-}2.02 & \phantom{0}19.4 \\
drawing                   & \phantom{-}1.61 & \phantom{0}29.2 \\
hand drawn                & \phantom{-}1.42 & \phantom{0}20.2 \\
vector illustration       & \phantom{-}1.38 & \phantom{0}26.1 \\
tattoo                    & \phantom{-}1.27 & \phantom{0}12.2 \\
white vector illustration & \phantom{-}1.22 & \phantom{0}29.2 \\
illustration              & \phantom{-}1.09 & \phantom{0}20.5 \\
sketch                    & \phantom{-}1.02 & \phantom{0}16.2 \\
step by step              & \phantom{-}0.53 & \phantom{0}25.8 \\
digital art               & \phantom{-}0.31 & \phantom{0}23.2 \\
% white background          & \phantom{-}0.81 & \phantom{0}5.3 \\
% step by step              & \phantom{-}0.81 & \phantom{0}0.4 \\
% art selected              & \phantom{-}0.63 & \phantom{0}1.1 \\
% art                       & \phantom{-}0.59 & \phantom{0}0.6 \\
% digital art selected      & \phantom{-}0.50 & \phantom{0}1.1 \\
% pencil step               & \phantom{-}0.30 & \phantom{0}0.6 \\
% person                    & -0.36 & 25.1 \\
% vector                    & -0.41 & \phantom{0}0.6 \\
% step                      & -0.45 & \phantom{0}0.7 \\
% dog                       & -0.53 & 27.2 \\
% white                     & -1.53 & 10.1 \\
\bottomrule
\end{tabular}
\end{subtable}
\begin{subtable}{0.3\textwidth}
\centering\small
% \vspace{0.14in}
\caption{ImageNet-C snow (base acc.: 64.6)}
\vspace{-0.04in}
% \resizebox{\textwidth}{!}{% 
\begin{tabular}{lcc}
\toprule
\phantom{rain drops falling}
& Score & Acc. \\
\midrule
snow falling       & \phantom{-}3.05 & 27.9 \\
rain falling       & \phantom{-}2.58 & 0.9 \\
rain drops falling & \phantom{-}2.52 & 26.1 \\
rain drops         & \phantom{-}2.25 & 26.7 \\
rain               & \phantom{-}2.14 & 51.6 \\
snow               & \phantom{-}1.83 & 54.2 \\
water drops        & \phantom{-}1.52 & 32.3 \\
falling            & \phantom{-}1.33 & 27.5 \\
water              & \phantom{-}1.02 & 51.1 \\
day                & \phantom{-}0.53 & 67.9 \\
% car                & \phantom{-}0.39 & 68.0 \\
% person playing     & \phantom{-}0.39 & 64.0 \\
% person             & \phantom{-}0.20 & 64.6 \\
% playing            & \phantom{-}0.11 & 56.7 \\
% dog playing        & \phantom{-}0.09 & 54.7 \\
% man                & \phantom{-}0.03 & 44.5 \\
% dog                & -0.05 & 68.0 \\
% dogs               & -0.08 & 63.2 \\
% black              & -0.28 & 63.0 \\
% biological species & -0.30 & 85.3 \\
\bottomrule
\end{tabular}
\end{subtable}
\begin{subtable}{0.3\textwidth}
\centering\small
% \vspace{0.14in}
\caption{ImageNet-C frost (base acc.: 67.7)}
\vspace{-0.04in}
% \resizebox{\textwidth}{!}{% 
\begin{tabular}{lcc}
\toprule
\phantom{rain drops falling}
& Score & Acc. \\
\midrule
room               & \phantom{0}0.97 & 53.4 \\
glass              & \phantom{0}0.83 & 47.1 \\
window             & \phantom{0}0.81 & 55.9 \\
snow               & \phantom{0}0.70 & 70.8 \\
water              & \phantom{0}0.58 & 65.5 \\
person playing     & \phantom{0}0.52 & 65.3 \\
tree               & \phantom{0}0.39 & 72.2 \\
person             & \phantom{0}0.33 & 65.7 \\
dogs playing       & \phantom{0}0.31 & 50.0 \\
car                & \phantom{0}0.31 & 62.4 \\
% make               & \phantom{0}0.20 & 71.4 \\
% man                & \phantom{0}0.09 & 52.0 \\
% close              & \phantom{0}0.08 & 73.2 \\
% dog                & \phantom{0}0.00 & 70.7 \\
% dogs               & \phantom{0}-0.02 & 64.5 \\
% biological         & \phantom{0}-0.11 & 87.1 \\
% black              & \phantom{0}-0.14 & 68.8 \\
% biological species & \phantom{0}-0.17 & 87.1 \\
% white              & \phantom{0}-0.23 & 68.5 \\
% species            & \phantom{0}-0.53 & 87.0 \\
\bottomrule
\end{tabular}
\end{subtable}
\end{table*}

\begin{table*}[ht!]
\caption{Candidates of bias keywords for Dollar Street.}
\vspace{-0.05in}
\centering\small
\begin{subtable}{0.4\textwidth}
\centering\small
\caption{Wardrobe (base acc.: 60.7)}
\vspace{-0.04in}
% \resizebox{\textwidth}{!}{% 
\begin{tabular}{lcc}
\toprule
\phantom{toilet paper bag}
& Score & Acc. \\
\midrule
cave    & \phantom{-}1.83  & \phantom{0}0.0  \\
laundry & \phantom{-}1.05  & 33.3 \\
man     & \phantom{-}0.67  & \phantom{0}0.0  \\
pile    & \phantom{-}0.34  & 50.0 \\
sleeps  & \phantom{-}0.30  & \phantom{0}0.0  \\
living  & -0.01 & \phantom{0}0.0  \\
shed    & -0.48 & \phantom{0}0.0  \\
clothes & -0.99 & 72.7 \\
full    & -1.10 & 71.4 \\
room    & -1.22 & 58.3 \\
\bottomrule
\end{tabular}
\end{subtable}
\begin{subtable}{0.4\textwidth}
\centering\small
\caption{Stove (base acc.: 50.0)}
\vspace{-0.04in}
% \resizebox{\textwidth}{!}{% 
\begin{tabular}{lcc}
\toprule
\phantom{toilet paper bag}
& Score & Acc. \\
\midrule
burns     & \phantom{-}0.90  & \phantom{0}0.0  \\
fire      & \phantom{-}0.80  & \phantom{0}0.0  \\
fireplace & \phantom{-}0.05  & \phantom{0}0.0  \\
cat       & \phantom{-}0.04  & \phantom{0}0.0  \\
sits      & -0.17 & \phantom{0}0.0  \\
room      & -0.17 & \phantom{0}0.0  \\
small     & -0.51 & 50.0 \\
sink      & -0.62 & \phantom{0}0.0  \\
kitchen   & -1.60 & 50.0 \\
stove     & -1.64 & 61.5 \\
\bottomrule
\end{tabular}
\end{subtable}
\begin{subtable}{0.4\textwidth}
\centering\small
\vspace{0.14in}
\caption{Plate rack (base acc.: 24.3)}
\vspace{-0.04in}
% \resizebox{\textwidth}{!}{% 
\begin{tabular}{lcc}
\toprule
\phantom{toilet paper bag}
& Score & Acc. \\
\midrule
bucket     & \phantom{-}0.78  & \phantom{0}3.8  \\
water      & \phantom{-}0.78  & \phantom{0}3.0  \\
small      & \phantom{-}0.13  & 25.0 \\
sink       & \phantom{-}0.03  & 29.5 \\
food       & -0.02 & 11.8 \\
full       & -0.51 & 21.6 \\
laundry    & -0.52 & 22.2 \\
kitchen    & -1.13 & 32.3 \\
dishes     & -1.41 & 25.0 \\
collection & -1.42 & 21.1 \\
\bottomrule
\end{tabular}
\end{subtable}
\begin{subtable}{0.4\textwidth}
\centering\small
\vspace{0.14in}
\caption{Toillet seat (base acc.: 46.0)}
\vspace{-0.04in}
% \resizebox{\textwidth}{!}{% 
\begin{tabular}{lcc}
\toprule
\phantom{toilet paper bag}
& Score & Acc. \\
\midrule
hole     & \phantom{-}0.65  & \phantom{0}0.0  \\
house    & \phantom{-}0.10  & 62.3 \\
property & \phantom{-}0.04  & 80.0 \\
basement & -0.09 & 42.9 \\
man      & -0.17 & 42.9 \\
image    & -1.03 & 81.6 \\
small    & -1.03 & 23.5 \\
room     & -1.50 & 58.1 \\
bathroom & -3.50 & 59.6 \\
toilet   & -4.70 & 71.4 \\
\bottomrule
\end{tabular}
\end{subtable}
\end{table*}
\begin{table*}[ht!]
\caption{Candidates of bias keywords for ImageNet.}
\vspace{-0.05in}
\centering\small
\begin{subtable}{0.4\textwidth}
\centering\small
\captionsetup{justification=centering}
\caption{Ant (base acc.: 30.0)}
\vspace{-0.04in}
% \resizebox{\textwidth}{!}{% 
\begin{tabular}{lcc}
\toprule
\phantom{motherboard removed}
& Score & Acc. \\
\midrule
flowers & 1.08 & 14.7 \\
flower  & 1.03 & 20.9 \\
bee     & 0.99 & 12.9 \\
tree    & 0.86 & 19.1 \\
spider  & 0.78 & 29.5 \\
fly     & 0.75 & 24.2 \\
beetle  & 0.58 & 30.4 \\
leaf    & 0.32 & 27.3 \\
close   & 0.12 & 33.3 \\
black   & 0.10 & 18.1 \\
\bottomrule
\end{tabular}
\end{subtable}
\begin{subtable}{0.4\textwidth}
\centering\small
\captionsetup{justification=centering}
\caption{Horizontal bar (base acc.: 70.8)}
\vspace{-0.04in}
% \resizebox{\textwidth}{!}{% 
\begin{tabular}{lcc}
\toprule
\phantom{motherboard removed}
& Score & Acc. \\
\midrule
swings     & 7.01 & \phantom{0}6.3  \\
playground & 5.09 & \phantom{0}9.5  \\
park       & 4.63 & \phantom{0}3.6  \\
swing      & 4.31 & 12.5 \\
child      & 3.47 & 27.7 \\
plays      & 2.83 & 20.7 \\
girl       & 2.52 & 22.0 \\
playing    & 2.14 & \phantom{0}4.1  \\
person     & 1.35 & 65.2 \\
boy        & 1.10 & 20.0 \\
\bottomrule
\end{tabular}
\end{subtable}
\begin{subtable}{0.4\textwidth}
\vspace{0.14in}
\centering\small
\captionsetup{justification=centering}
\caption{Stethoscope (base acc.: 69.1)}
\vspace{-0.04in}
% \resizebox{\textwidth}{!}{% 
\begin{tabular}{lcc}
\toprule
\phantom{motherboard removed}
& Score & Acc. \\
\midrule
baby        & \phantom{-}1.23  & 24.4 \\
boy         & \phantom{-}1.23  & 28.0 \\
girl        & \phantom{-}0.71  & 36.2 \\
person      & \phantom{-}0.51  & 36.2 \\
student     & \phantom{-}0.44  & 30.4 \\
nurse       & \phantom{-}0.01  & 72.9 \\
doctor      & -0.81 & 88.4 \\
hospital    & -0.87 & 56.1 \\
medical     & -0.99 & 88.3 \\
stethoscope & -3.04 & 93.3 \\
\bottomrule
\end{tabular}
\end{subtable}
\begin{subtable}{0.4\textwidth}
\vspace{0.14in}
\centering\small
\captionsetup{justification=centering}
\caption{Monastery (base acc.: 53.0)}
\vspace{-0.04in}
% \resizebox{\textwidth}{!}{% 
\begin{tabular}{lcc}
\toprule
\phantom{motherboard removed}
& Score & Acc. \\
\midrule
interior  & \phantom{-}1.12  & 17.6 \\
built     & \phantom{-}0.53  & 54.2 \\
cathedral & \phantom{-}0.35  & 36.7 \\
person    & \phantom{-}0.29  & 60.5 \\
century   & \phantom{-}0.06  & 58.8 \\
city      & \phantom{-}0.03  & 56.7 \\
church    & -0.01 & 53.3 \\
temple    & -0.16 & 46.9 \\
courtyard & -0.50 & 58.5 \\
town      & -0.64 & 60.6 \\
\bottomrule
\end{tabular}
\end{subtable}
\end{table*}

\end{appendices}

\end{document}